\documentclass[12pt]{iopart}
\usepackage{iopams}

\usepackage{natbib}
\usepackage{har2nat}
\setcitestyle{aysep={,}}

\usepackage{gensymb}
\usepackage{graphicx}
\usepackage{multirow}
\usepackage{xcolor}
\usepackage{hyperref}
\hypersetup{colorlinks=true,citecolor=[rgb]{0.08 0.54 0.80},linkcolor=[rgb]{0.08 0.54 0.80},urlcolor=[rgb]{0.08 0.54 0.80},hypertexnames=false}

\newcommand{\eqref}[1]{(\ref{#1})}
\newcommand{\Fig}{Figure}
\newcommand{\fig}{figure}

\begin{document}

\title[Manuscript for \PMB]{Fast and Automatic Periacetabular Osteotomy Fragment Pose Estimation Using Intraoperatively Implanted Fiducials and Single-View Fluoroscopy}

\author{R~B~Grupp\textsuperscript{1}, R~J~Murphy\textsuperscript{2}, R~A~Hegeman\textsuperscript{3}, C~P~Alexander\textsuperscript{4}, M~Unberath\textsuperscript{1}, Y~Otake\textsuperscript{5}, B~A~McArthur\textsuperscript{6,7}, M~Armand\textsuperscript{3,4,8}\\and R~H~Taylor\textsuperscript{1}}

\address{\textsuperscript{1}Department of Computer Science, Johns Hopkins University, Baltimore, MD, USA}
\address{\textsuperscript{2}Auris Health, Inc., Redwood City, CA, USA}
\address{\textsuperscript{3}Research and Exploratory Development Department, Johns Hopkins University Applied Physics Laboratory, Laurel, MD, USA}
\address{\textsuperscript{4}Department of Orthopaedic Surgery, Johns Hopkins Medicine, Baltimore, MD, USA}
\address{\textsuperscript{5}Graduate school of Information Science, Nara Institute of Science and Technology, Ikoma, Nara, Japan}
\address{\textsuperscript{6}Department of Surgery and Perioperative Care, Dell Medical School, University of Texas, Austin, TX, USA}
\address{\textsuperscript{7}Texas Orthopedics, Austin, TX, USA}
\address{\textsuperscript{8}Department of Mechanical Engineering, Johns Hopkins University, Baltimore, MD, USA}
\ead{grupp@jhu.edu}
\vspace{10pt}
\begin{indented}
\item[]June 2020
\end{indented}

\begin{abstract}
Accurate and consistent mental interpretation of fluoroscopy to determine the position and orientation of acetabular bone fragments in 3D space is difficult.
We propose a computer assisted approach that uses a single fluoroscopic view and quickly reports the pose of an acetabular fragment without any user input or initialization.
%
Intraoperatively, but prior to any osteotomies, two constellations of metallic ball-bearings (BBs) are injected into the wing of a patient's ilium and lateral superior pubic ramus.
One constellation is located on the expected acetabular fragment, and the other is located on the remaining, larger, pelvis fragment.
The 3D locations of each BB are reconstructed using three fluoroscopic views and 2D/3D registrations to a preoperative CT scan of the pelvis.
The relative pose of the fragment is established by estimating the movement of the two BB constellations using a single fluoroscopic view taken after osteotomy and fragment relocation.
BB detection and inter-view correspondences are automatically computed throughout the processing pipeline.
%
The proposed method was evaluated on a multitude of fluoroscopic images collected from six cadaveric surgeries performed bilaterally on three specimens.
Mean fragment rotation error was $2.4 \pm 1.0$ degrees, mean translation error was $2.1 \pm 0.6$ mm, and mean 3D lateral center edge angle error was $1.0 \pm 0.5$ degrees. The average runtime of the single-view pose estimation was $0.7 \pm 0.2$ seconds.
%
The proposed method demonstrates accuracy similar to other state of the art systems which require optical tracking systems or multiple-view 2D/3D registrations with manual input.
The errors reported on fragment poses and lateral center edge angles are within the margins required for accurate intraoperative evaluation of femoral head coverage.
\end{abstract}
\footnotesize
%
\vspace{2pc}

\vspace{1pc}
\noindent This submission includes a supplementary document. An additional video with detailed descriptions of the methods and example results is available online at:~\href{https://youtu.be/0E0U9G81q8g}{https://youtu.be/0E0U9G81q8g}.
\normalsize
%
%
%
%
%
\section{Introduction}\label{sec:intro}
\begin{figure}
\begin{indented}
\item[]
\begin{center}
\includegraphics[width=0.75\linewidth]{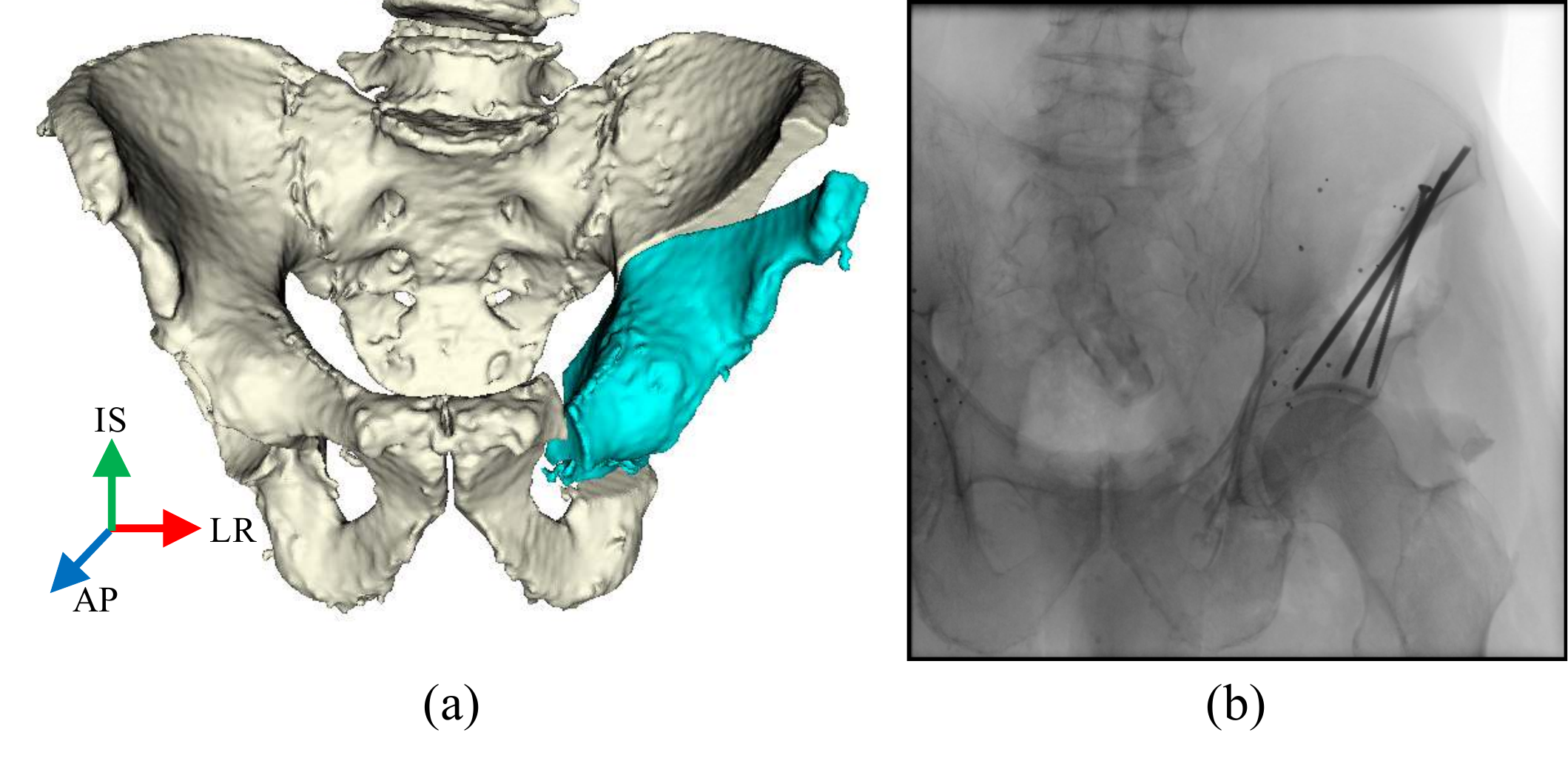}
\end{center}
\end{indented}
\caption{Examples of an adjusted acetabular fragment visualized in 3D (a) and in a corresponding 2D fluoroscopic image (b).
The fragment pose shown in (a) was estimated using the view shown in (b).
A precise model of the acetabular fragment is not required by the proposed method; the 3D bone surfaces in (a) were constructed using a preoperative plan of the osteotomies.
The anatomical axes of the anterior pelvic plane are also shown in (a); left/right (LR) as X-axis, inferior/superior (IS) as Y-axis, and anterior/posterior (AP) as Z-axis.}
\label{fig:example_xray_and_pose}
\end{figure}

Patients suffering from developmental dysplasia of the hip (DDH) typically have severe pain and reduced coverage of the femoral head, which can lead to joint osteoarthritis and subluxation of the femur~\cite{gala2016hip}.
Joint-preserving pelvic osteotomies, such as the Periacetabular osteotomy (PAO), treat DDH by reorienting the hip joint for increased femoral head coverage~\cite{ganz1988new}.
Specifically for PAO, four osteotomies are performed about the acetabulum, fracturing it from the pelvis and allowing it to be adjusted to the desired pose~\cite{ganz1988new}.
In the conventional approach, PAO surgeons rely on 2D X-ray images, tactile feedback, experience and acumen to navigate the surgery~\cite{ganz1988new}.
Clinicians typically assess femoral head coverage intraoperatively using specific radiographic measurements, such as the lateral center edge (LCE) angle~\cite{wiberg1939studies}, derived from fluoroscopy~\cite{troelsen2009surgical}.
However, this approach does not indicate the full 3D alignment of the acetabular fragment, nor does it describe additional biomechanical parameters, which have the potential to improve surgical outcomes~\cite{hipp1999planning,armand2005outcome,armiger2009three,niknafs2013biomechanical,liu2016evaluation}.
A 3D example of a relocated fragment and a corresponding 2D fluoroscopic view is shown in \fig~\ref{fig:example_xray_and_pose}.

In this paper, we propose a processing pipeline that is capable of automatically reporting fragment poses from a single fluoroscopic view with mean runtimes below one second.
The pipeline is inspired by Roentgen stereometric analysis (RSA) techniques, which use metallic ball-bearings (BBs) to track the movement of bones or surgical implants over time~\cite{selvik1990roentgen}.

Two constellations of BBs are injected into the patient's pelvis prior to osteotomy: one co-located on the expected acetabular bone fragment and the other on the larger pelvis portion.
The 3D locations of the BBs are reconstructed using three fluoroscopic views of the constellations.
Once the acetabulum is relocated, the 3D orientation and position of the fragment is automatically calculated using a single fluoroscopic view.

Existing navigation systems for PAO have traditionally relied on optical tracking devices for tool and bone fragment tracking~\cite{langlotz1997first,akiyama2010computed,murphy2015development,liu2016periacetabular,takao2017comparison}.
Due to their limited fields of view and susceptibility to line-of-sight obstructions, optical trackers introduce significant positioning constraints which are challenging to satisfy in already crowded operating rooms.
Since our method only requires a small BB injection device and fluoroscopy, which is already common throughout orthopaedic operating rooms for joint surgery,
we believe the proposed fluoroscopic method is more easily deployable than approaches relying on optical tracking technology.
Furthermore, the registration process with an optical tracker requires a certain amount of bone exposure and may become more challenging when using minimally invasive incisions~\cite{troelsen2008new}.
Since hip surgeries incorporating RSA do not demonstrate significantly different operating times than corresponding surgeries without RSA~\cite{shah2018routine},
the process of injecting BBs into the pelvis should not substantially interfere with existing surgical workflows.
Compared to existing approaches which leverage fluoroscopy~\cite{grupp2019pose}, our method only requires a single fluoroscopic image per pose estimate, does not rely on any knowledge of the 3D fragment shape, and runs without user initialization in a fraction of the time.

After BB injection, the proposed method does not require any specialized equipment or additional workflow.
Moreover, the pose estimation executes quickly and automatically between fluoroscopic captures. 
The primary clinical contribution of this paper is the ability to report 3D orientation and position of the acetabular fragment, while requiring minimal modification to an existing surgical workflow.
In terms of technical contribution, this paper is the first method leveraging intraoperatively constructed fiducial constellations to automatically recover point correspondences and poses of multiple objects moving non-coherently in uncalibrated single-view fluoroscopy.
\subsection{Related Work}\label{sec:rel_work}
Early navigation systems for PAO, and other pelvic osteotomies, relied on optical trackers~\cite{langlotz1997first,langlotz1998computer,mayman2002kingston,akiyama2010computed}, or custom cutting guides~\cite{radermacher1998computer,otsuki2013developing}.
These systems only provided intraoperative assistance during performance of the acetabular osteotomies; pose estimates of the relocated fragment were not produced.

More recent systems have focused on reporting the pose of a relocated fragment~\cite{murphy2015development,liu2016periacetabular,murphy2016clinical,takao2017comparison,de2018reliability,grupp2019pose}.
Fragment pose updates may be provided in real-time by directly attaching an optically tracked rigid body to the fragment as demonstrated in~\cite{liu2016periacetabular}.
However, attaching a large rigid body to the acetabular region is challenging, especially when using a minimally invasive technique specialized for PAO~\cite{troelsen2008new}.
In order to estimate fragment poses and avoid the attachment of an extra rigid body, \cite{murphy2015development} and~\cite{takao2017comparison} digitize specific points on the fragment with an optically tracked pointer tool after each adjustment of the fragment.
This digitization adds minor overhead to the operative time in~\cite{murphy2015development} and causes some ambiguity between rotation and translation in~\cite{takao2017comparison}.
In~\cite{murphy2015development}, fragment pose errors ranged from $1.4-1.8\degree$ in rotation and $1.0-2.2$ mm in translation.

Eliminating the need for optical tracking systems, \cite{grupp2019pose} used multiple fluoroscopic views to track the acetabular fragment, ipsilateral femur, and pelvis.
A multiple-component intensity-based 2D/3D registration of patient anatomy was used, requiring no external objects and maintaining compatibility with any PAO approach.
However, the method suffers from several limitations and constraints that interfere with a typical surgical workflow:
\begin{itemize}
	\item An approximate AP fluoroscopic view, two additional views, and manual annotation of a single anatomical landmark are required to initialize the method
	\item Accuracy of the approach degrades as intraoperative fragment shapes differ from preoperatively planned shapes
	\item The computation time on state-of-the-art hardware is not real-time, approximately $25$ seconds.
\end{itemize}
To overcome these limitations, the methods described in this paper leverage implanted BBs and extend RSA-related techniques to automatically track the migration of the acetabular fragment using a single view per adjustment.

Since its introduction in the 1970's, RSA has been used for a variety of applications~\cite{selvik1990roentgen}, including the longitudinal analysis of orthopaedic implant migration~\cite{valstar2002use}, bone growth~\cite{karrholm1984changes}, and even PAO stability~\cite{mechlenburg2007safe}.
Recent work in the RSA community has incorporated 2D/3D registration technology to track the movement of bones and implants without relying on inserted BBs~\cite{de2008image,seehaus2012markerless}.
Similar to~\cite{grupp2019pose}, these methods require manual input and multiple X-ray views.

The implantation of BBs for intraoperative fragment tracking during PAO was first introduced in~\cite{murphy2013computer} and~\cite{armand2018biomechanical}.
However, this approach is not easily incorporated into a surgical workflow, since it requires: the manual identification of BB correspondences, multiple post-osteotomy views, and a calibrated CBCT C-arm.

Several methods for automatic 3D BB reconstruction have been developed by the CBCT community~\cite{hamming2009automatic,yaniv2009localizing,dang2012robust,choi2014fiducial}.
These methods require a calibrated C-arm, leverage more than three projections, or rely on an orbital motion constraint to help establish correspondences.
For intraoperative coronary artery reconstruction, epipolar constraints have been applied to automatically prune invalid point correspondences between two~\cite{yang2009novel}, and three~\cite{blondel2006reconstruction}, fluoroscopic views.
Structure-from-Motion pipelines operate in a similar fashion and use the dense correspondences found in large photographic collections to reconstruct rigid structures in 3D~\cite{snavely2008modeling,westoby2012structure,schonberger2016structure}.

Methods using a known 3D marker constellation, two 2D X-ray views, and varying levels of manual interaction have been developed for patient positioning and motion compensation in radiation therapy~\cite{schweikard2000robotic,litzenberg2002daily,aubry2004measurements}.

Single plane RSA was proposed in order to avoid the less-common, bi-planar, imaging devices used in RSA~\cite{yuan2002roentgen}.
However, the method requires the use of a calibration cage, does not address the establishment of 2D/3D correspondences, and was only evaluated for single object pose recovery.

Automatic pose and correspondence estimation between a single rigid collection of BBs and one fluoroscopic view was described in~\cite{tang2000fiducial} and~\cite{kang2013robustness}.
The poses of custom tailored fiducial objects using lines and ellipses may also be computed automatically in a single view~\cite{jain2005ftrac,steger2013marker}.
These poses are generally restricted to tracking the relative motion of the C-arm, since the relationship between the patient's anatomy and the intraoperatively inserted fiducial is typically unknown.

The method of~\cite{tang2000fiducial} was incorporated into fluoroscopic systems for estimating the poses of multiple BB constellations required for knee kinematics~\cite{tang2004accurate,ioppolo2007validation}.
However, the mechanism used to identify constellation membership and establish 2D/3D correspondences, was not described.

The pipeline proposed in this paper is able to accurately, quickly, and automatically provide pose estimates of a relocated bone fragment during PAO.
No reliance on external tracking devices is required.
Furthermore, the pose estimation method does not require: a calibrated C-arm, multiple-views, a specific constellation pattern, accurate knowledge of the fragment shape, or any manual establishment of correspondence.
\section{Materials and Methods}\label{sec:methods}
\begin{figure}
\begin{indented}
\item[]
\begin{center}
\includegraphics[width=0.578\linewidth]{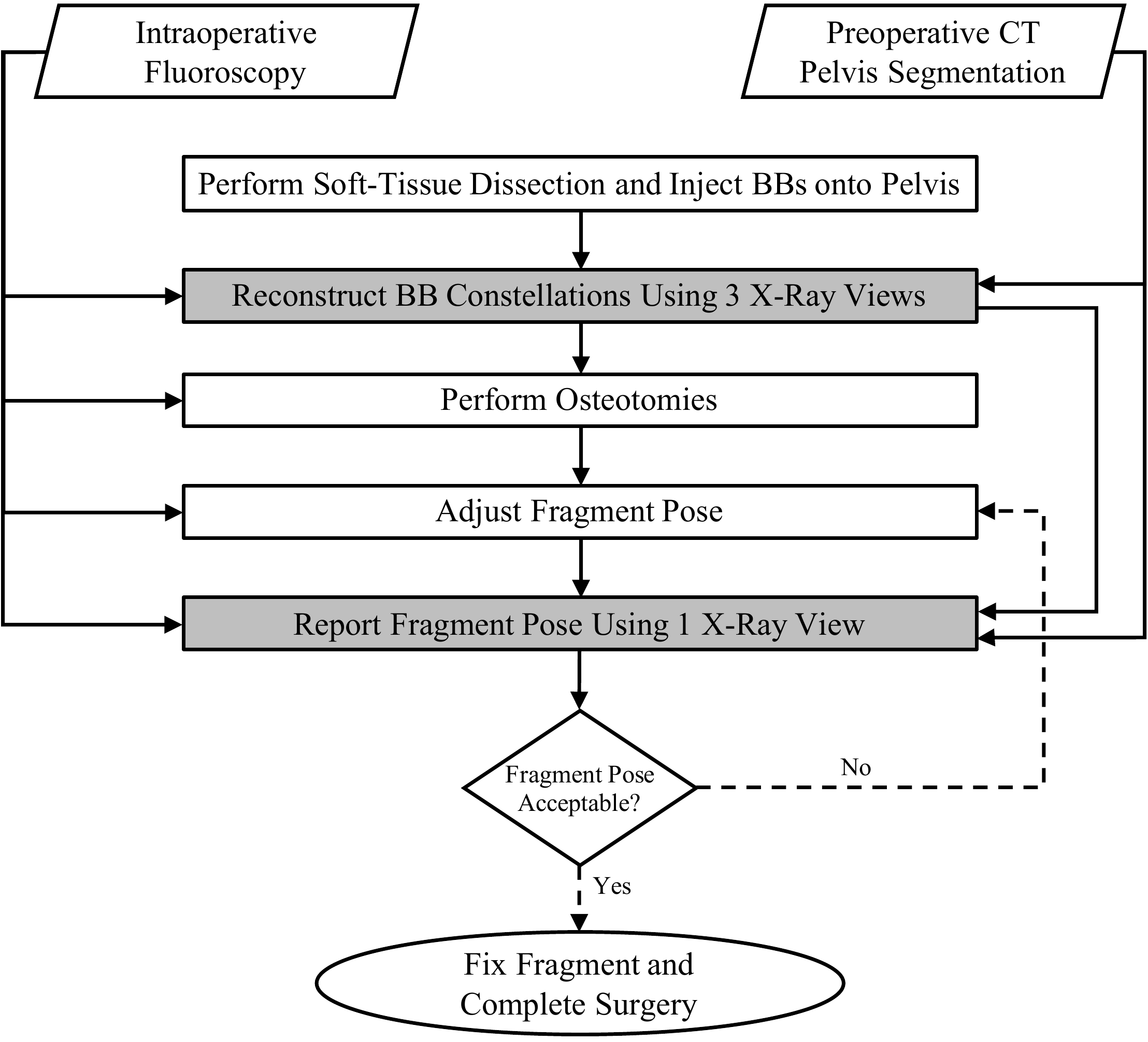}
\end{center}
\end{indented}
\caption{A summary of the surgical workflow proposed for this method, including the data required for each step.
			 The key contributions of this work are the BB reconstruction and single-view pose estimation components, which are highlighted in gray.
			 Detailed workflows for the reconstruction and pose estimation components are shown in supplementary figures~\ref{fig:bb_recon_workflow} and~\ref{fig:pose_est_workflow_combined}, respectively.}
\label{fig:workflow}
\end{figure}

%
The method introduced in this paper requires some preoperative processing and two distinct phases during the surgery.
CT scanning, segmentation of the anatomy, and anatomical landmark digitization make up the preoperative processing.
The first intraoperative phase is performed only once and consists of BB injection and reconstruction.
Pose estimation of the acetabular fragment from a single fluoroscopic view represents the second intraoperative addition.
In order to achieve the desired amount of femoral head coverage, it is typical for a surgeon to iterate between collecting fluoroscopy and adjusting the fragment.
Therefore, our processing combines intelligent pruning and GPU acceleration to avoid any significant delay to the workflow.
\Fig~\ref{fig:workflow} shows the workflow of the proposed method at a high level.
Detailed flow charts of the reconstruction and pose estimation sub-components are provided in supplementary section~\ref{sec:supp_methods}.
Full details of the preoperative processing, intraoperative BB reconstruction, and intraoperative fragment pose estimation are now provided.
\subsection{Preoperative Processing}\label{sec:preop_proc}
Preoperative processing proceeds identically to that in~\cite{grupp2019pose}, which we briefly describe here.
A lower torso CT scan is obtained and resampled to have 1 mm isotropic voxel spacing.
An automated method~\cite{krvcah2011fully} is used for an initial segmentation of the pelvis and femurs; any inconsistencies around the femoral head and acetabulum are cleaned up manually.
Anatomical landmarks are manually annotated to define the anterior pelvic plane (APP) coordinate system~\cite{nikou2000description}, and also for later use as initialization of pre-osteotomy pelvis registrations.
The origin of the APP is set at the center of the ipsilateral femoral head, and the mapping from APP coordinates to the CT volume coordinates is denoted by $T_{APP}^{V}$.
Six additional landmarks are manually annotated in order to create a planned fragment shape, which is only used to visualize the intraoperative movement of the fragment.
Examples of the APP axes orientation and a planned fragment shape are shown in \fig~\ref{fig:example_xray_and_pose}.
\subsection{Intraoperative BB Reconstruction}\label{sec:intraop_recon}
A Halifax Biomedical Inc. injection device is used to implant two, four-BB constellations onto the ipsilateral side of the patient's pelvis, with one constellation lying on the area expected to lie on the acetabular fragment and the other on the larger pelvis fragment.
BB injection is conducted after performing soft-tissue dissection, but prior to osteotomy.
The injection device may be maneuvered either with direct sight or fluoroscopic guidance to ensure that the BB constellations will lie on on separate bone fragments after performance of the osteotomies.
For the experiments conducted in this paper, fluoroscopy was not used during the BB insertion process.
Although at least three BBs must not be colinear within each constellation in order to later track their rigid movement~\cite{west2001fiducial}, this constraint is not difficult to satisfy in practice due to the 3D curvature of the bone.

Three distinct fluoroscopic views are collected while the patient anatomy remains stationary.
For each view, a variation of the radial symmetry algorithm~\cite{loy2003fast} is used to automatically locate each BB in 2D.
Parameter values and details of this approach are listed in supplementary section~\ref{sec:supp_methods_rad_symmetry}.
The 3D locations of each BB are constructed by recovering the relative pose information of each view, establishing inter-view BB correspondences, and performing triangulation~\cite{hartley2003multiple}.
\begin{figure}
\begin{indented}
\item[]
\begin{center}
\includegraphics[width=0.75\linewidth]{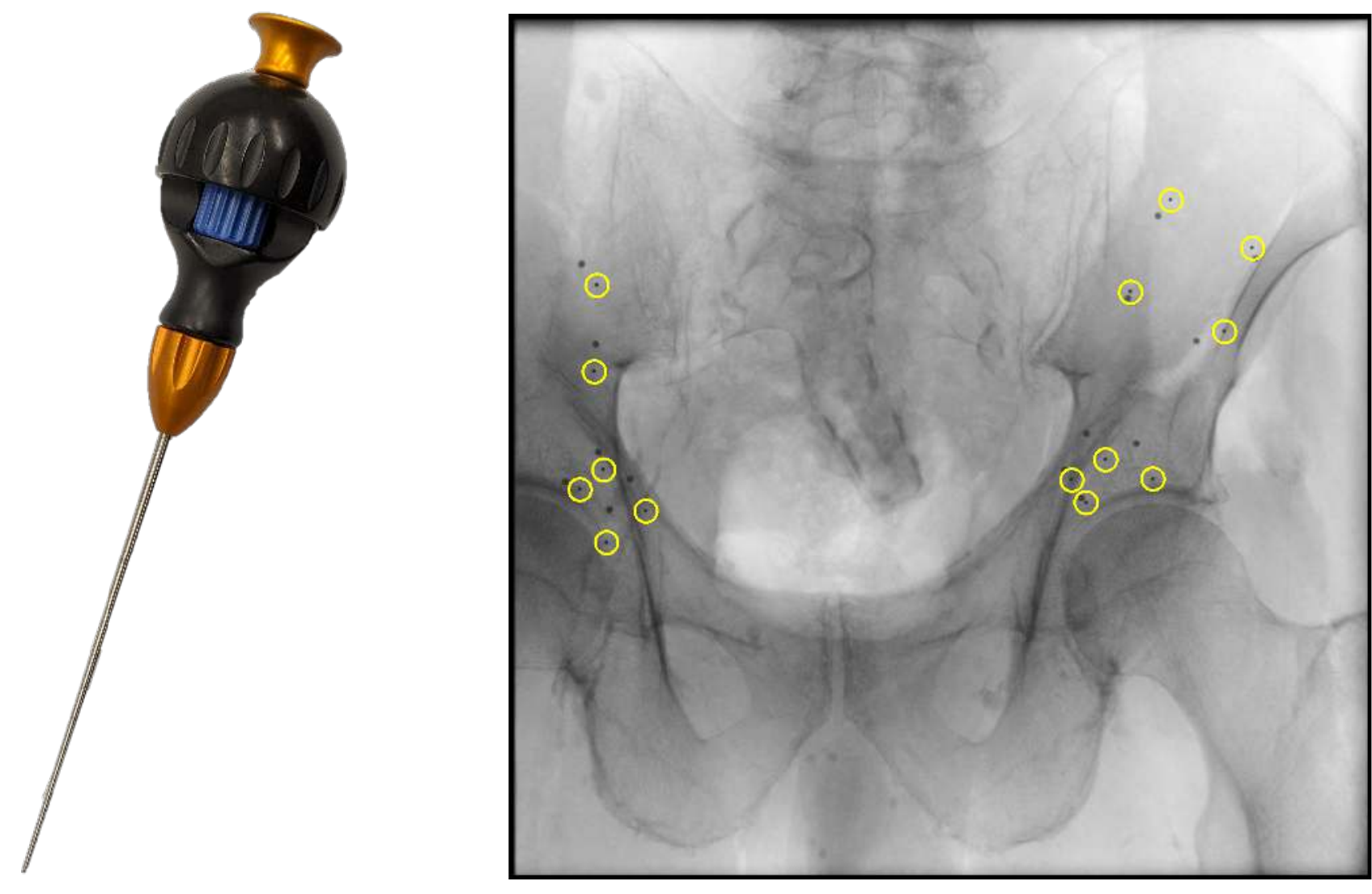}
\end{center}
\end{indented}
\caption{The Halifax bead injection device used in four of the cadaver surgeries is shown on the left.
On the right, a pre-osteotomy fluoroscopic image is shown with automatic detections of injected beads highlighted by yellow circles; every injected BB was detected.
The larger BBs were used to help establish the ground truth pose of the fragment and as such, are not used and not detected during intraoperative pose estimation.}
\label{fig:injector_and_precut_xray}
\end{figure}

Using the strategy laid out in~\cite{grupp2019pose}, relative poses between the three views are recovered by performing 2D/3D rigid registrations of the patient's preoperative pelvis to each view.
Since some of our pre-osteotomy views have excessive pelvic tilt and violate the approximate AP view assumption, we select more than the single landmark described in~\cite{grupp2019pose} for initialization of the pipeline.
Supplementary section~\ref{sec:app_intensity_regi} describes the parameters used for the intensity-based registrations.

%
%
\begin{figure}
\begin{indented}
\item[]
\begin{center}
\includegraphics[width=0.75\linewidth]{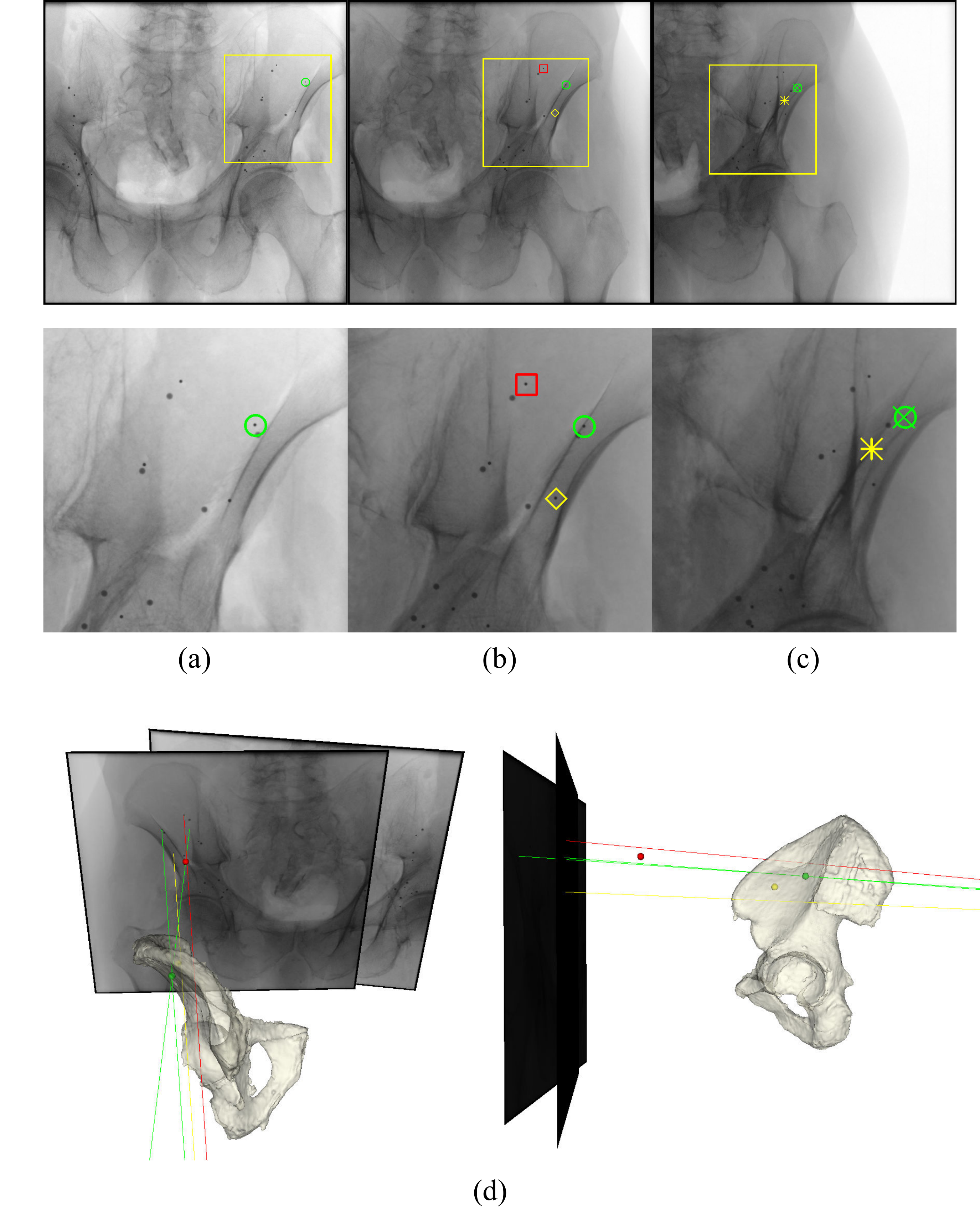}
\end{center}
\end{indented}
\caption{A visual example of the pre-osteotomy reconstruction process for a \textit{single} BB.
Three fluoroscopic views used for reconstruction are shown in (a), (b), and (c).
The initial two-view triangulations are derived from (a) and (b), while (c) is used for re-projections of initial triangulations.
Regions pertinent to this example are indicated by yellow boxes, and are magnified in the bottom row.
3D renderings of the patient's ipsilateral hemi-pelvis and the relative location of the C-arm detector for the first two views are shown in (d).
The green circle in (a) indicates the location of a detected BB, whose 3D location is to be reconstructed.
In (b), the green circle shows the detected BB location with true correspondence to BB in (a); the red square and yellow diamond show detected locations with incorrect correspondence.
In (d), the three colored spheres are initial triangulations of the BB from (a) when matched with the BBs of varying colors in (b).
The red sphere is not located on the pelvis and its candidate correspondence is pruned.
However, the green and yellow spheres are located on the pelvis and must be checked using (c).
Lines between the X-ray source and BB locations on the detector are colored consistently with (a), (b), and (c); note the intersection between the green lines.
The green circle in (c) indicates the detected location of the BB in true correspondence with the green circles in (a) and (b).
The green ``X'' is the re-projection of the green sphere from (d) and the yellow asterisk is the re-projection of the yellow sphere.
Since the green sphere was triangulated using a correct correspondence, its re-projected distance to the BB detection in (c) is very small compared to the re-projected distance of the yellow sphere, which was triangulated using an incorrect correspondence.}
\label{fig:recon_example}
\end{figure}
BB correspondences are automatically established using a combination of anatomical information and the multiple-view geometry between the three C-arm poses.
Two of the views are selected to create a candidate set of two-view, single-BB, correspondences and triangulated 3D points.
Although we have made no assumptions about the geometry of these views, one of the views was always an approximate AP orientation with a variable amount of pelvic tilt.
The candidate correspondences are created by first considering all possible combinations of single-BB correspondences between the two views, and pruning candidates that result in a triangulated point located more than 10 mm away from the pelvis surface.
The red sphere shown in \fig~\ref{fig:recon_example} is an example of a correspondence pruned in this way.
Candidate three-view correspondences are constructed by pairing each of the remaining two-view correspondences with every 2D BB detection in the third view.
For each candidate three-view correspondence, the two-view 3D triangulation is re-projected into the third view and the distance to the hypothesized 2D match is recorded.
Intuitively, re-projection distances for valid correspondences should be smaller than distances from invalid matches, as shown with the green and yellow re-projections in \fig~\ref{fig:recon_example}.
Correct three-view correspondences are established by greedily selecting the candidate correspondences with minimum re-projection distances in the third view.
The final 3D reconstructions are triangulated using the correct three-view correspondences.
In this way, the third view is used to enforce consistency and refine the 3D triangulation.
A visual example is shown in \fig~\ref{fig:recon_example} and a more formal description is located in supplementary section~\ref{sec:app_bb_recon}.

After pruning reconstructed BBs on the contralateral side, the process is completed by classifying the remaining BBs as fragment/non-fragment using a $K$-Means clustering of the BB positions ($K=2$).

Annotation speed and computation time is not a critical factor at this point in the procedure, since the BB constellations are not required until the fragment has been relocated; osteotomies may be performed immediately after the three fluoroscopic views are obtained.
\subsection{Intraoperative Pose Estimation}\label{sec:intraop_pose}
After osteotomies have been performed and the acetabular fragment has been relocated, a single fluoroscopic image may be used to recover the fragment's pose with respect to the APP, $\Delta_{APP}$.
Once the the poses of the ilium and fragment BB constellations with respect to the C-arm, $T_C^{IL}$ and $T_C^{FR}$, are recovered, $\Delta_{APP}$ is computed as in \eqref{eq:frag_pose_app}.
\begin{equation} \label{eq:frag_pose_app}
	\Delta_{APP} = T_V^{APP} T_C^{FR} T_{IL}^C T_{APP}^{V}
\end{equation}
Since the BB constellations are constructed in the original pelvis volume coordinate frame, the composition of $T_C^{FR} T_{IL}^C$ is valid and maps points on the preoperative fragment region to their adjusted locations.
Using $\Delta_{APP}$, the current pose of the fragment may be visualized (\fig~\ref{fig:example_xray_and_pose}), and pose parameters or biomechanical (e.g. LCE) angles may also be displayed.

As was the case for each fluoroscopic view used for pre-osteotomy BB reconstruction, the radial symmetry algorithm is used to detect each BB in the 2D fluoroscopic image automatically.
Since the 3D/2D BB correspondences are not yet established, it is not feasible to directly apply classic PnP approaches~\cite{hartley2003multiple} for calculation of $T_C^{IL}$ or $T_C^{FR}$.
Since manual identification is tedious and error-prone, an automatic method to establish correspondences is the appropriate intraoperative strategy.

One possible, although na{\"i}ve, approach would be to enumerate over all possible correspondences and their poses.
Digitally Reconstructed Radiographs (DRRs) and similarities with the fluoroscopic image would be computed for each candidate pose, with the actual pose implied by the best similarity score.
When exactly all 8 BBs are detected in the view and 4 correspondences are used to establish a candidate pose, state-of-the-art GPUs may be capable of efficiently computing the 1,680 similarity scores of all possible ilium BB correspondences.
However, the tractability of performing a brute-force search over similarity scores quickly diminishes as additional BBs are detected.
Additional detections may result from false detections on the screws and K-wires used to fix the fragment, and are possibly exacerbated during bilateral cases due to detections on the contralateral side.
Examples of 2D BB detections in fluoroscopy images are shown in supplementary
figures~\ref{fig:spec1_fluoro},~\ref{fig:spec2_fluoro}, and~\ref{fig:spec3_fluoro},
with the number of BB detections per image varying between 7 and 21.
For specimen~1, there were 13, 12, and 21 total detections in projections 1, 2, and 3, yielding 17,160, 11,880, and 143,640 possible ilium BB correspondences, respectively.
Even with state-of-the-art GPUs, the sheer number of possible poses precludes the brute-force strategy from working in an intraoperatively compatible timeframe.
However, we shall describe a procedure for pruning the number of candidate poses by several orders of magnitude, enabling the required similarity scores to be intraoperatively computed through GPU acceleration.

\subsubsection{General Pose Pruning Strategy}\label{sec:meth_general_pose_pruning}
For a given 4-BB constellation, the general pruning strategy enumerates over each 3-BB sub-constellation.
Furthermore, the full set of possible 3-BB 3D/2D correspondences for each sub-constellation is examined.
Potential solutions to the P3P problem~\cite{fischler1981random} are considered for each set of correspondences.
Since we are concerned with pose estimation using fluoroscopic imagery, our approach to the P3P problem assumes that the BB constellation lies between the X-ray source and detector.
This assumption enables solutions to be ignored which: are impossible given the rigid structure of the constellation, or which place the BB constellation too close to the X-ray source.
Many poses that would be produced from incorrect 3-BB correspondences are discarded in this way.
In addition to the point sets and hypothesized correspondences, a set of source-to-detector ratios is also required as input to the P3P solver.
The source-to-detector ratios are used to back-project one of the 2D BB detections to possible 3D locations, simplifying the pruning problem.
Full details of this approach are described in supplementary section~\ref{sec:app_p3p}.
Solutions reported by the P3P solver are further pruned according to anatomical constraints.
The candidate source-to-detector ratios and anatomical constraints differ for the ilium and fragment BB constellations.
\subsubsection{Ilium BB Pose Estimation}\label{sec:meth_il_bb_pose_est}
\begin{figure}
\begin{indented}
\item[]
\includegraphics[width=\linewidth]{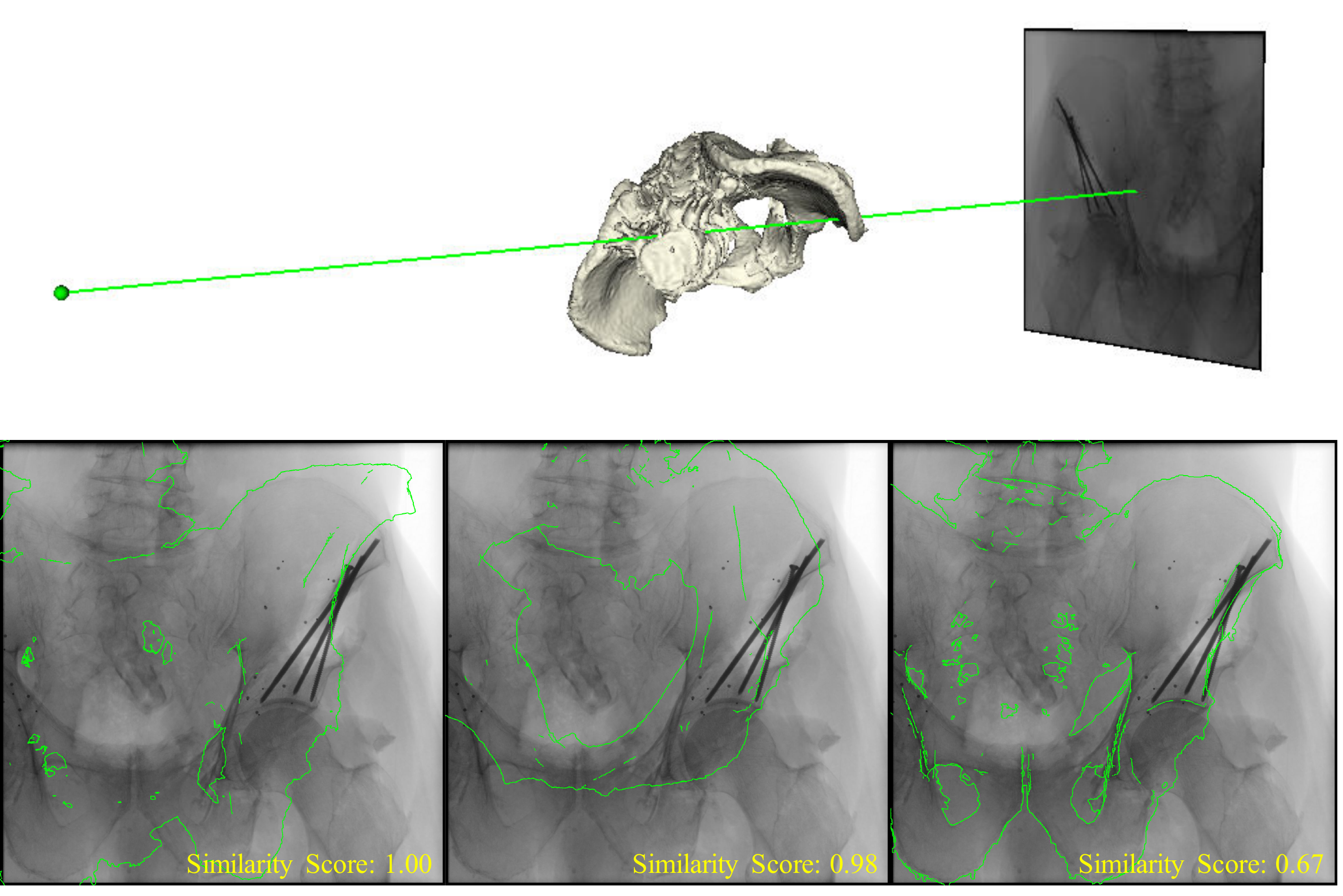}
\end{indented}
\caption{The top row shows an ilium pose pruned for excessive difference from the reference AP orientation ($137\degree$ about the AP axis).
The green sphere indicates the X-ray source with a green line connecting to the principal point on the X-ray detector.
For this example, the candidate correspondences were able to satisfy the constraints of the P3P solver.
However, the implausibility of the pose reveals the incorrectness of the correspondences.
The bottom row depicts several examples of ilium poses and correspondences used for initialization of the full-pelvis intensity-based, 2D/3D, registration.
Green edges, derived from a specific pelvis pose, are overlaid over the intraoperative fluoroscopic image.
Agreement between the overlaid edges and base image indicates agreement between the hypothesized pose and true pose.
Image similarity scores are listed in the bottom right of each overlay.
The scores are computed from DRRs, computed at each candidate pose, and the intraoperative fluoroscopic image.
Lower scores indicate better similarity, with the bottom right example representing the most likely pose of the four.}
\label{fig:example_bad_il_prune}
\end{figure}

The pose of the ilium BB constellation is recovered first and is then used to assist with establishing the pose of the fragment BB constellation.
A set of 129 uniformly spaced source-to-detector ratios is used for each ilium P3P invocation: $\left \{ 0.6 + 0.003125k | k = 0, 1, \dots, 128 \right \}$.
Using the APP coordinate frame, a reference AP orientation of the pre-osteotomy pelvis, with respect to the C-arm, is constructed and used for pruning anatomically implausible ilium poses.
The AP orientation has the following properties: the patient is supine with the X-ray detector placed anteriorly, the AP axis is parallel to the C-arm depth axis, the IS axis is parallel to the 2D image row axis with the top of the image more superior than the bottom, and the LR axis is parallel to the 2D image column axis.
Each candidate P3P pose is examined to obtain the difference in orientation from the reference AP pose and an Euler decomposition is used to obtain rotation angles about each anatomical axis.
Poses are pruned when the magnitude of any Euler angle is greater than $60\degree$.
Using such a large range of allowable angles permits all reasonable C-arm orientations while eliminating highly unlikely poses, such as those that place the detector beneath, or nearly orthogonal with, the surface of the operating table.
An example of a pose pruned using this logic is shown in the top row of \fig~\ref{fig:example_bad_il_prune}.
Using each remaining candidate ilium pose, the original fragment BB constellation is projected into the view; e.g.\ where the fragment BBs would be located in 2D had the fragment not been moved.
Since the majority of fragment movement consists of rotation, the re-projected fragment BBs should lie nearby to 2D BB detections.
Poses are pruned when less than 3 of the fragment BBs are projected inside the bounds of the 2D image.
For each projected fragment BB, the distance to the nearest 2D detection is calculated, and the three BBs with the smallest nearest distances are recorded.
When the mean distance associated with these three BBs is greater than 200 pixels the candidate ilium pose is pruned.
%
%

Once the general pruning strategy has been completed for the ilium BB constellation, the na{\"i}ve brute-force, intensity-based, approach is used to select the best ilium pose from the remaining candidates.
The bottom row of \fig~\ref{fig:example_bad_il_prune} shows several examples of image similarity calculated from poses derived from different correspondences.
This pose is used as initialization for an intensity-based 2D/3D registration of the pre-osteotomy pelvis to the fluoroscopic image.
Details of the intensity-based registration parameters are listed in supplementary section~\ref{sec:app_intensity_regi}.
Using the pose estimate computed during the intensity-based registration, final ilium BB correspondences are established by re-projecting the 3D ilium BBs into 2D.
Correspondences are greedily assigned based on the minimum 2D distances between projected BB locations and detected 2D BB locations.
However, no correspondence is established for projected BBs with minimum distances greater than $10.5$ pixels.
When less than two correspondences are established we consider the algorithm to have failed in establishing the ilium pose and no further processing is performed.
The ilium pose is set to the intensity-based registration pose when exactly two correspondences are established.
When three or four correspondences are established, the ilium pose is refined by optimizing over the corresponding ilium BB re-projection distances starting from the intensity-based pose as the initial guess.

The set of 2D BB detections is pruned down to exclude: BBs already matched to the ilium, and any BBs that are distant from the expected location of the fragment.
A BB is considered distant if the closest, re-projected, fragment BB is greater than 200 pixels away.
This is a variation of the process previously used for pruning ilium poses by re-projection of 3D fragment BBs.

\subsubsection{Fragment BB Pose Estimation}\label{sec:meth_frag_bb_pose_est}
\begin{figure}
\begin{indented}
\item[]
\begin{center}
\includegraphics[width=0.75\linewidth]{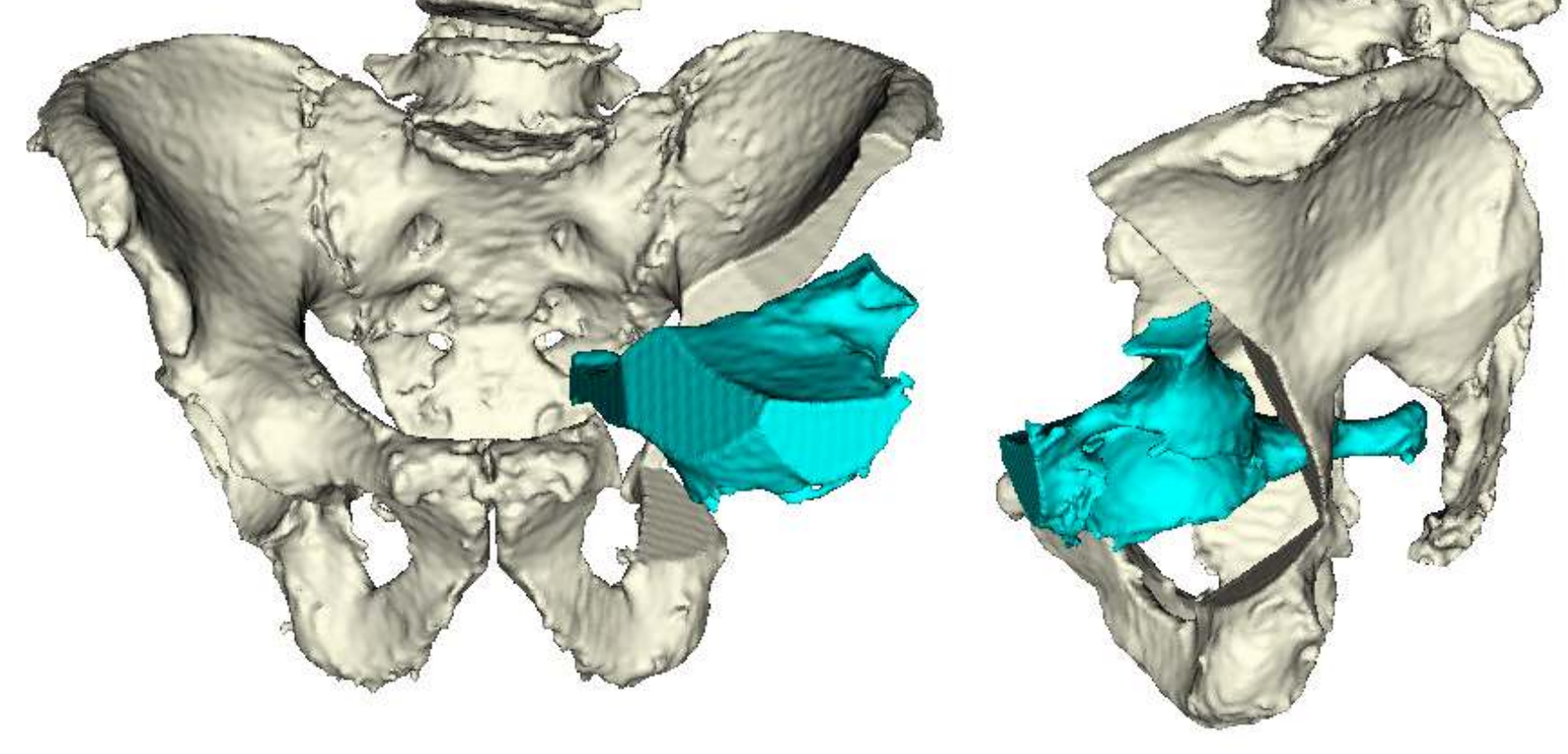}
\end{center}
\end{indented}
\caption{An example of an implausible fragment pose which was pruned due to a large rotation of $142\degree$.
Despite their incorrectness, the candidate correspondences used to compute this candidate fragment pose were able to satisfy the P3P solver constraints.}
\label{fig:example_bad_frag_constraint_prune}
\end{figure}

The fragment pose recovery is started by conducting the general pruning strategy over candidate fragment BB correspondences and poses.
Since approximate depth of the BBs is known from the ilium pose recovery process, only 33 source-to-detector ratios are passed to the P3P solver.
A reference source-to-detector ratio, $\hat{r}$, is computed by mapping the centroid of the fragment 3D BB constellation into the C-arm coordinate frame using the ilium pose.
The source-to-detector ratios are then uniformly sampled about this reference: $\left \{ \hat{r} \pm 0.003125k | k = 0, 1, \dots, 16 \right \}$.
Using each solution produced by the P3P solver, the relative pose of the fragment is computed using \eqref{eq:frag_pose_app}.
 Any relative pose with rotation magnitude greater than $60\degree$ or translation magnitude greater than 30 mm is pruned.
\Fig~\ref{fig:example_bad_frag_constraint_prune} shows an example fragment pose that was pruned in this fashion.

Due to the difficult nature of the chiseling process, the true shape of the acetabular fragment usually differs from the preoperatively planned shape.
For this reason, image similarities are not used to select the best candidate returned from the general pruning process.
Instead, the best candidate is selected by choosing the pose yielding the largest number of matching BBs and the smallest mean re-projection distance.
The match criterion used for ilium matches is reused here.
When less than 3 BBs are matched, the method reports failure.
However, the fragment pose is refined by an optimization over re-projection distances if at least 3 BBs are matched.
The optimization is regularized by the translation magnitude of the fragment pose relative to the APP.
This regularization is reasonable, since the fragment movement is believed to consist primarily of rotation and the approximate depth is known from the ilium pose.

It is important to note that the approach described here only requires correspondences to be established for three ilium BBs and three fragment BBs.
Therefore, the proposed method provides some robustness to occlusion, since it is unlikely that more than one BB from a single constellation will be occluded for any given view.
Likewise, it is still feasible to obtain fragment pose estimates when a single BB (per constellation) becomes dislodged from the bone.
\subsection{Cadaver Experiments}\label{sec:meth_cad_experiments}
Surgeries performed on three, non-dysplastic, cadaveric specimens were used to evaluate the proposed method.
Specimens~1, 2, and 3 were aged~89, 87, 94 and were male, female, and male, respectively.
Preoperative processing and planning was performed bilaterally for each specimen and six PAOs were performed by our surgeon co-author, B.A.M.
The Halifax injector, using BBs of 1 mm diameter, was only used during surgeries for specimens~2 and~3.
For specimen~1, bone burs were created on the surface of the pelvis, and $1.5$ mm diameter BBs were affixed with cyanoacrylate.
The larger BBs were also inserted into specimens~2 and~3, but were only used for ground truth calculations.
For both BB insertion approaches, direct visual guidance was used to ensure a sufficient separation of the two BB constellations.
A comprehensive discussion of BB insertion and the fragment pose ground truth protocol is found in~\cite{grupp2019pose}.
Ground truth poses for specimen~1 were calculated using the 2D/3D known BB constellation approach, whereas specimens~2 and~3 use the 3D/3D method.
For specimen~1, the 3D BB positions were manually digitized from a postoperative CT scan and mapped into the preoperative CT frame using a rigid registration between the two images.
Pre-osteotomy and post-osteotomy poses of the BB constellations were recovered using a series of 2D/3D, paired point, registrations between the 3D BBs and manually annotated 2D BB locations in the fluoroscopic views.
The ground truth fragment pose was defined as the rigid movement of the fragment BBs between the pre-ostetomy and post-osteotomy poses.
In addition to the pre and postoperative CT scans for specimens~2 and~3, a CT scan was also collected after BB implantation, but prior to any osteotomies.
The positions of the pre and post-osteotomy BBs were manually identified in the corresponding CT scans.
A paired point 3D/3D rigid registration between the pre and post-osteotomy BBs was used to calculate the ground truth pose of the fragment.

Three fluoroscopic views were used to reconstruct the pre-osteotomy 3D BB constellations for each surgery.
Pose estimation of the relocated fragment was conducted on 3 separate fluoroscopic images, each with different viewing geometries.
All fluoroscopy was obtained using a Siemens CIOS Fusion C-arm with 30 inch flat panel detector.
The intrinsic parameters of the C-arm were naively approximated using metadata present in DICOM images produced by the system.
A source-to-detector distance of 1020 mm, isotropic pixel spacings of $0.194$ mm, no skew or distortion, and a principal point at the center of the projection were used.
For C-arms with minimal distortion, previous work has demonstrated that approximate intrinsics yield accurate reconstructions of small objects and accurate relative pose estimates~\cite{jain2005c}.
The approximate intrinsics used in this paper are therefore reasonable, as the reconstructions are of small BBs and the final relative fragment pose estimates are not with respect to the C-arm coordinate frame.

A video example of the full pipeline, from preoperative planning to intraoperative pose estimation is available online at: \href{https://youtu.be/0E0U9G81q8g}{https://youtu.be/0E0U9G81q8g}.
\section{Results}\label{sec:results}
\subsection{Intraoperative BB Reconstruction}\label{sec:results_recon}
For pre-osteotomy BB reconstruction, there were no missed detections in the 2D images and a single false detection in one image.
Table~\ref{tab:results_bb_recon_errors} summarizes the reconstruction errors.
The mean reconstruction error of the larger BBs implanted into specimen~1 was $2.6$~mm.
For specimens 2 and 3, the mean reconstruction error of the smaller, injected, BBs was $1.4$~mm.
The mean computation time for the entire reconstruction pipeline was $8.3 \pm 0.4$~seconds.
When excluding the 2D/3D full pelvis registration time required for relative pose recovery of the C-arm, the BB detection, correspondence establishment and reconstruction took $0.7 \pm 0.1$ seconds.
Timing measurements were conducted using a single NVIDIA P100 (PCIe) GPU and seven cores of an Intel Xeon E5-2680 v4 CPU.
A detailed breakdown of these runtimes is found in supplementary table~\ref{tab:supp_results_bb_recon_times}.
\begin{table}
\caption{A summary of BB reconstruction errors for each surgery, specified by the cadaver specimen number and operative side.
The means and standard deviations of reconstruction errors are given for the separate ilium and fragment BB constellations and also the entire set of BBs.
For each surgery, four BBs were reconstructed for each of the ilium and fragment constellations.}
\label{tab:results_bb_recon_errors}
\begin{indented}
\item[]
\begin{tabular}{@{}l l l l}
\br
Surgery & \multicolumn{3}{c}{Reconstruction Errors (mm)} \\ \cline{2-4}
& Ilium BBs & Fragment BBs & All BBs  \\
\mr
1 Left   & $2.5 \pm	    0.4$ &	$3.2  \pm	0.2$ &	$2.9  \pm	0.5$ \\
1 Right & $2.1 \pm	0.2$ &	$2.7  \pm	0.5$ &	$2.4  \pm	0.5$ \\
2 Left   & $1.6  \pm	0.3$ &	$1.4  \pm	0.2$ &	$1.5  \pm	0.3$ \\
2 Right & $1.3  \pm	0.5$ &	$1.1  \pm	0.1$ &	$1.2  \pm	0.3$ \\
3 Left   & $1.3  \pm	0.3$ &	$1.1  \pm	0.4$ &	$1.2  \pm	0.3$ \\
3 Right & $1.4  \pm	0.2$ &	$1.6  \pm	0.2$ &	$1.5  \pm	0.2$ \\
\br
\end{tabular}
\end{indented}
\end{table}
%

Thumbnails of each fluoroscopy image used for BB reconstruction during the cadaver experiments are found in supplementary
figures~\ref{fig:spec1_fluoro_precut},~\ref{fig:spec2_fluoro_precut}, and~\ref{fig:spec3_fluoro_precut}.
\subsection{Intraoperative Pose Estimation}\label{sec:results_pose_est}
On average, 99.6\% of the maximum number of \textit{ilium} poses are pruned by the P3P solver step.
Using anatomical constraints, the remaining poses are pruned by an average of 95.9\%.
After this pruning, an average of 57 candidate ilium poses were used for the exhaustive image similarity searches.
The maximum number of \textit{fragment} poses were pruned by an average of 97.3\% using the P3P solver, and anatomical constraints pruned 81.2\% of the remaining poses on average.
In order to determine the final fragment BB correspondences, an average of 20 fragment poses remained to be searched after pruning.
A summary of the maximum number of ilium and fragment poses considered, and the actual poses considered due to pruning, is shown in supplementary table~\ref{tab:supp_results_pose_pruning}.

\begin{figure}
\begin{indented}
\item[]
\begin{center}
\includegraphics[width=\linewidth]{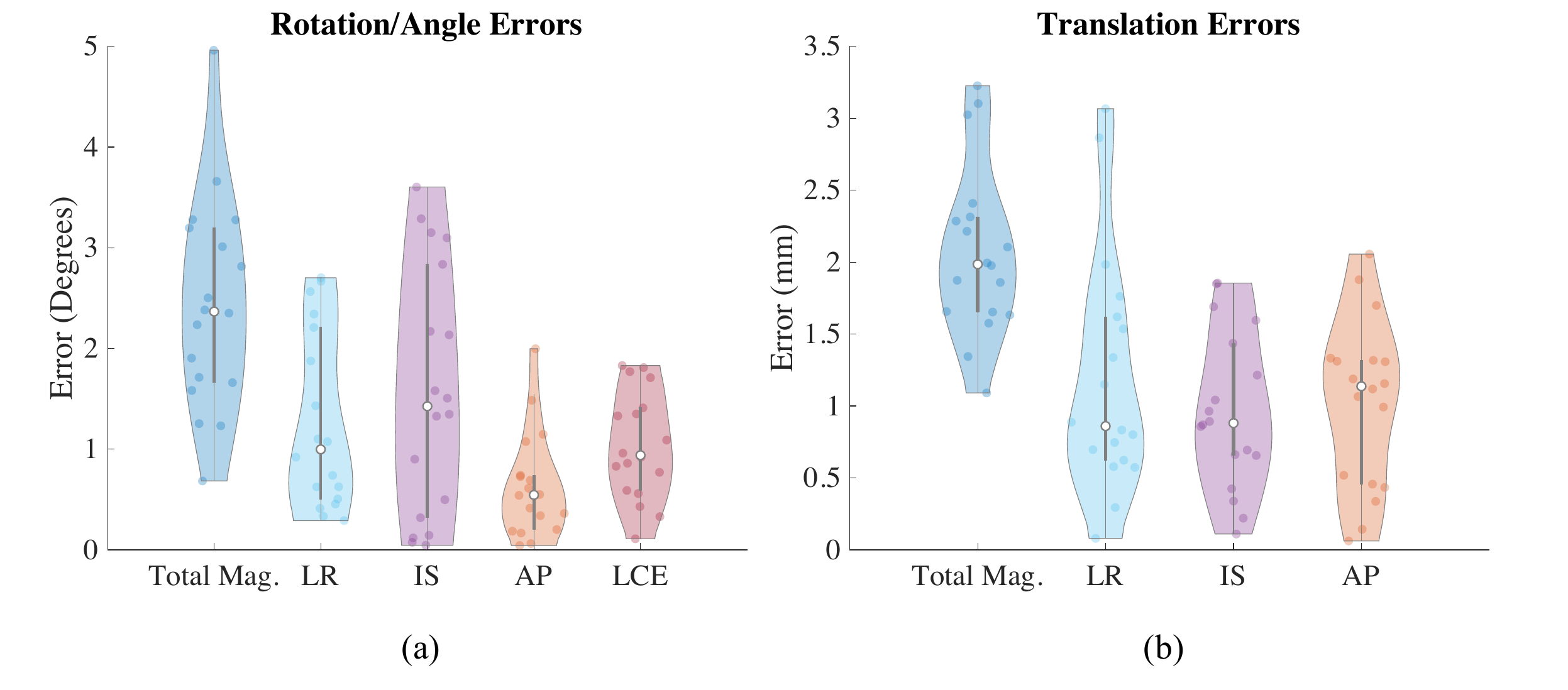}
\end{center}
\end{indented}
\caption{A visualization of the distributions of errors associated with each fragment pose estimate.
Rotation and lateral center edge (LCE) angle errors are shown in (a), and translation errors are shown in (b).
For rotation and translation errors, the magnitude is listed along with the errors about the left-right (LR), inferior-superior (IS), and anterior-posterior (AP) anatomical axes.
Each circle represents the error of a pose estimate, the white circle indicates the median error, the vertical bar indicates the $[25,75]$ percentile range, and a kernel density function is fit around the samples.}
\label{fig:results_pose_est_violins}
\end{figure}

Pose estimation was successfully performed on 18 total views (3 per surgery).
The distributions of fragment rotation, LCE angle, and translation errors is shown in \fig~\ref{fig:results_pose_est_violins}.
Errors in rotation were below $3\degree$ for 12 of the 18 cases, with a mean of $2.4\degree$.
When the rotation errors were decomposed about anatomical axes, only rotation about the IS axis had errors greater than $3\degree$.
In terms of both mean and standard deviation, rotation measurements about the AP axis were the most accurate, followed by LR, and then IS.
The maximum 3D LCE angle error was $1.8\degree$ and the mean was $1.0\degree$.
The mean translation error was 2.1 mm, and was less than 3 mm for 15 of the 18 estimates.
Mean translation errors about the anatomical axes were all within 0.2 mm of each other, and the maximum difference between standard deviations was 0.3 mm.
An entire listing of errors for each pose estimate is shown in supplementary table~\ref{tab:supp_results_pose_errors}.

All four ilium BBs were matched in 6 of the 18 cases and all four fragment BBs were matched in 16 of the 18 cases. 
The mean rotation, translation, and LCE angle errors for estimates with 4 ilium BBs matched were $1.7\degree$, $2.1$ mm, and $1.0\degree$, respectively.
With less than 4 ilium BBs matched, the mean errors were $2.8\degree$, $2.0$ mm, and $1.1\degree$, respectively.
With 4 fragment BBs matched, the mean rotation, translation, and LCE angle errors were $2.3\degree$, $2.1$ mm, and $1.0\degree$, respectively.
The mean errors were $3.1\degree$, $1.5$ mm, and $1.6\degree$, when less than 4 fragment BBs matched.
Supplementary table~\ref{tab:supp_results_bb_detects} includes a full summary of the number of BB detections and matches in each image.

In the third view for the left side of specimen~1, one ilium BB was outside the image bounds and not detected.
On the left side of specimen~2, one of the ilium BBs was occluded by K-wire in each view and therefore not detected.
Analysis of the postoperative CT revealed that this BB was actually dislodged by either: performance of the ilium osteotomy or insertion of the K-wire.
The missed ilium detections in views 1 and 2 on the right side of specimen~2, were occluded by screws.
Occlusion by K-wire also caused the missed ilium detection in view 2 on the right side of specimen~3.
However, according to the postoperative CT this ilium BB was also displaced from the bone.
The missed fragment BB detections were caused by K-wire occlusion.
There were no missed detections of ipsilateral BBs that were present in the scene and not occluded by screws or K-wires.

False detections corresponding to contralateral BBs were observed in 8 of the projections, with an average of 5 detections over these projections.
An average of 4 false BB detections were caused by the screws visible in the 6 projections used for specimen~1.
No other false BB detections were reported.

The mean computation time for the single-view pose estimation was $0.7 \pm 0.2$ seconds, and was measured using the same hardware used to record reconstruction times.
Supplementary table~\ref{tab:supp_results_bb_detects_times_pose_est} provides a detailed breakdown of these runtimes.

Thumbnails of each fluoroscopy image used for fragment pose estimation during the cadaver experiments are found in supplementary figures~\ref{fig:spec1_fluoro},~\ref{fig:spec2_fluoro}, and~\ref{fig:spec3_fluoro}.
\section{Discussion}\label{sec:discussion}
Although one third of fragment pose estimates had rotation errors larger than $3\degree$, LCE angle errors were well below the $3\degree$ success criteria identified in~\cite{grupp2019pose}.
This indicates that the proposed method is able to quantify the amount of lateral femoral head coverage, resulting from an intraoperatively relocated acetabulum, within clinically acceptable error thresholds.
The $3\degree$ LCE angle error threshold was chosen so that clinicians can target a correction with a $35\degree$ LCE angle and have confidence that the actual LCE lies in a $30\degree - 40\degree$ range, which is indicative of dysplasia correction and associated with lower incidents of failure~\cite{hartig2012factors}.

Given the automatic nature of the method and the relatively quick runtime, it should be feasible for clinicians to smoothly move between making pose adjustments to the fragment, taking fluoroscopic shots, and receiving feedback regarding the current pose estimate.

The mean rotation error when less than the full number of BBs were matched in either constellation, was $1.1\degree$ greater than the mean rotation error over the cases matching full BB constellations.
However, mean translation and mean LCE angle error were less effected by unmatched ilium BBs.
When only 3 fragment BBs were matched, the mean LCE angle error was $0.6\degree$ larger than the mean LCE angle error associated with all fragment BBs matched.
Therefore, the number of matched BBs in each constellation may be used to convey confidences in the estimated poses.
When less than 4 fragment BBs are matched, confidence in any rotation and LCE angle would be lowered.
For cases when all 4 fragment BBs were matched, but less than 4 ilium BBs were matched, confidence in LCE angle would remain unaffected, however confidence in general rotation would be reduced.

Highlighting the robustness of the method, all LCE errors remained below the $3\degree$ error threshold, even when BBs were missing from the view or occluded.
The method was also robust against BBs which were displaced from the pelvis, but remained in the field of view and detected.
This was demonstrated on the right side of specimen~3, where an ilium BB had been dislodged from the bone and was detected in views 1 and 2.
Since the P3P solver does not return solutions for non-rigid transformations of the constellations, the displaced BB was not matched, despite the detection of all four ilium BBs in these views.

At least three BBs from each constellation must be detected in the fluoroscopy images for pose estimation to succeed.
If fewer than three BBs were automatically detected, the system could be operated using locations manually annotated by a human operator.
Although this would eliminate the possibility of real-time pose estimation, the system could still be used to verify fragment poses after a sequence of manually guided adjustments.
Since complexity of pose estimation processing is dependent on the number of BBs detected, a large number of false-positive detections may also preclude real-time application of the proposed approach.
These exceptional cases were not encountered during the experiments of this paper.

Average BB reconstruction errors for specimen~1 were greater than those of specimens~2 and~3.
This was most likely caused by larger 2D BB localization errors for specimen~1.
This is to be expected, since the BBs used for specimen 1 were larger than those used for specimens~2 and~3.
Only one of the six LCE errors for specimen~1 was greater than $1\degree$, indicating that the proposed method is not dependent on a single size of BBs.

As part of future work, we plan to also incorporate additional biomechanical indicators such as the anterior center edge angle and the acetabular index.
These measurements will enable a more complete intraoperative indication of femoral head coverage, which could potentially improve surgical outcomes.


The performance of the method compares favorably to the fiducial-free method (FFM) proposed in~\cite{grupp2019pose}.
When a postoperatively segmented fragment shape was retrospectively used for pose estimation, the FFM was reported to have mean rotation error of $2.2\degree$, mean translation error of 2.2 mm, and mean LCE error of $1.1\degree$.
The mean rotation error of the proposed method is only slightly larger than the FFM mean rotation error, while translation and LCE angle errors of the proposed method are slightly smaller.
When the FFM uses an intraoperatively refined version of a preoperatively planned fragment shape, the mean rotation, translation, and LCE angle errors increase to $3.5\degree$, 2.5 mm, and $1.8\degree$, respectively.
Considering that an accurate segmentation of the fragment shape is not available intraoperatively, the method proposed in this paper provides a more accurate assessment of fragment pose and femoral head coverage than the FFM.

In the current state of practice, preoperative CT data cannot be effectively used for intraoperative assessment of anatomical angles.
As a result, contemporary preoperative imaging usually consists solely of standing radiographs.
Although the proposed method requires a preoperative CT of the patient to be collected, the patient specific CT may be replaced with a statistical atlas of pelvis anatomy~\cite{otake2015comparison} in the future.
In this approach, the patient's anatomy would be reconstructed using a deformable 2D/3D registration between patient-specific 2D X-ray images and the atlas~\cite{sadowsky2007deformable,hurvitz2008registration,zheng2010statistical,kang2016simultaneous}.
A precise cartilage model is required for a comprehensive biomechanical analysis, including estimates of the joint contact pressure~\cite{armiger2009three}.
Since a statistical atlas may not be capable of satisfactorily reconstructing the cartilage model, partial CT of the patient's acetabulum may be used to augment the statistical model~\cite{chintalapani2010statistical,grupp2016pelvis}.
Furthermore, the process of collecting standing radiographs and a partial CT scan of the pelvis could potentially deliver a much smaller effective radiation dose than a complete CT scan of the pelvis.

The only manual portion of the intraoperative pipeline is the annotation of anatomical landmarks during BB reconstruction.
Recent advances in fluoroscopic deep learning enable these landmarks to be localized automatically~\cite{bier2019learning}, and incorporated into a completely automatic intraoperative pipeline, further reducing the impact on existing surgical workflows~\cite{grupp2019automatic}.
It is possible that an extension of these approaches for 2D BB localization could be faster, and more robust to other metallic objects, than the current radial symmetry approach.
The reconstruction and pose estimation processing pipelines do not rely on specific details of the 2D BB detection method, and improved BB detection algorithms may be substituted as necessary.

Although the registration framework leveraged from~\cite{grupp2019pose} is a highly optimized C++ library with OpenCL GPU acceleration, the pruning algorithms described in this paper were implemented as serial C++ routines.
We believe faster fragment pose estimation times should be possible, since the candidate poses and correspondences evaluated in each pruning phase are not dependent on one another, and the computations may be done in parallel.
For the current number of BBs proposed in this paper (8), pre-osteotomy BB reconstruction runtimes are dominated by pelvis registration times.
Any efforts to speed up reconstructions of a similar number of BBs should therefore focus on improving the 2D/3D registration pipeline.
However, the ``brute force'' approach used to construct all possible inter-view correspondences should also be improved if the number of BBs or point fiducials were to substantially increase.
Some efficiency could be gained by using epipolar constraints to prune implausible correspondences prior to the initial triangulation step.

Although optical tracking systems theoretically enable the continuous and real-time tracking of objects, this is not actually achieved by the optical tracker-based systems which require the repeated digitization of points lying on the fragment~\cite{murphy2015development,murphy2016clinical,takao2017comparison,de2018reliability}.
This digitization process is not real-time in nature and potentially slower than collecting another fluoroscopic image followed by the pose estimation computation proposed in this paper.

Since surgical equipment could potentially dislodge implanted fiducial objects during the acts of chiseling or drilling, we believe that intraoperative tool tracking is an essential component of future work.
A navigational capability for the osteotomes and drill should allow surgeons to avoid collisions with fiducial objects.
Real-time pose estimates of tools could be obtained by augmenting patterns of BBs to each object, collecting fluoroscopic views containing both tool and pelvis BBs, and applying methods similar to those proposed in this paper.

The risks associated with fiducial dislodgment may be further reduced by not using BBs and instead affixing \textit{temporary} structures to the bone during surgery and removing the structures once the fragment is satisfactorily fixed in place.
Sufficiently large structures should be chosen so that the object may be easily removed if it were to become dislodged.

For example, two deformable metallic grids of wires could be impressed on the bone surfaces of the fragment and ilium.
Wire intersection points would be treated as point fiducials and sharp feet on the underside would facilitate insertion into bone.
Each grid could be pressed against the surface of the pelvis, temporarily attached, and removed at the conclusion of the surgery.
If an osteotome or K-wire were to come into contact with the grid during chiseling or drilling, the grid would most likely deform and become partially detached from the pelvis.
However, removal of this larger grid should be significantly easier and lower risk, compared to the removal of a small, loose, BB.

Sets of percutaneously inserted wires with radiographic markers located at the tips of each wire could also serve as temporarily affixed fiducial objects.
Prior to osteotomy, sets of three or four wires would be percutaneously inserted into both the ilium and fragment bone regions.
As each wire tip is expected to remain fixed throughout the surgery, the radiographic markers could be used as point fiducials.

For either of these proposed fiducial objects, future work will include the development of a new 2D point fiducial detection method, which would replace the radial symmetry algorithm used in this paper.
However, all other components of the reconstruction and pose estimation pipelines would remain unchanged, as they only rely on the list of point fiducial locations produced by the radial symmetry implementation.

Assuming a sufficient navigational capability to avoid intraoperative BB dislodgment, long-term BB implantation should be clinically viable.
This is indicated by several long-term studies which rely on RSA beads implanted into the pelvis in order to measure the stability of acetabular THA implants~\cite{nieuwenhuijse2012good,klerken2015high,otten2016stability}.
To the best of our knowledge no complications have arisen when RSA beads have become detached from the bone~\cite{lawrie2003insertion} and are not at risk of moving into the joint space~\cite{shah2018routine}.
Since the methods described in this paper are not currently intended for long-term analysis of the joint, it may also be feasible to use biodegradable implanted fiducial markers in place of standard tantalum beads~\cite{stallmann2005biodegradable}.

When the proposed method is used alongside workflows which already rely on fluoroscopy for validating the fragment's pose, no substantial additional radiation dose is expected to be delivered to patients.
The three additional views required for 3D reconstruction of the implanted BBs do not represent a particular risk and will certainly impart a smaller dose than a CBCT collection.
Since the proposed method is capable of reporting fragment pose estimates for a wide range of view geometries, the traditional process of collecting many fluoroscopic views needed to obtain an appropriate view for human interpretation should be eliminated.
Therefore, we anticipate that the system could actually reduce the number of fluoroscopic views used for checking the fragment pose.
The radiation dose delivered to surgeons and operating team members when using fluoroscopy during PAO has been found to be safe so long as proper protective equipment was used~\cite{mechlenburg2009radiation}.
Further studies, under realistic clinical conditions, need to be performed in order to accurately determine the expected radiation doses imparted by the proposed system.

The proposed pose estimation workflow may also be used for cadaveric PAO training.
Pose estimates provided by the system could act as feedback for the mental estimates of the surgeon.
In this way, the system may improve surgeons' association of tactile sensing and fluoroscopic interpretation with a fragment's true pose.

As none of the cadaveric specimens exhibited hip dysplasia, no postoperative assessment of clinical adequacy was conducted.
The primary focus of these experiments was to determine the feasibility, accuracy and runtime of the proposed pose estimation algorithm and workflow.
Cadaveric specimens with uncorrected hip dysplasia are not readily available as a result of the advanced ages typically associated with the specimens.
Therefore, we believe that further studies on dysplastic hips could potentially be the focus of a clinical trial.
\section{Conclusion}\label{sec:conclusion}
This paper has proposed a new method for pose estimation of acetabular fragments using fluoroscopy and two constellations of intraoperatively implanted BBs.
Cadaveric studies have shown that the method is able to provide clinically accurate estimates of the LCE angle, a well-established indicator of femoral head coverage.
Once the BB constellations have been reconstructed in 3D, all fragment poses are calculated automatically using a single-view, and in sub-second runtime.
No other surgical equipment beyond a flat panel C-arm and BB injector is required.
The C-arm does not need to be calibrated, encoded, or motorized.
Unlike other fluoroscopic approaches, accurate knowledge of the bone fragment's shape is not necessary.
For these reasons, the proposed method provides minimal deviation from the standard surgical workflow, and should be easily mastered by clinicians already performing RSA.
\ack
We are grateful for the assistance of Mr. Demetries Boston during the cadaveric surgeries.
This research was supported by NIH/NIBIB grants R01EB006839, R21EB020113, MEXT/JSPS KAKENHI 26108004, Johns Hopkins University Internal Funds, and a Johns Hopkins University Applied Physics Laboratory Graduate Student Fellowship.
Part of this research project was conducted using computational resources at the Maryland Advanced Research Computing Center (MARCC).
Finally, we would also like to thank the anonymous reviewers for their insightful comments which helped to improve the quality of this manuscript.
\section*{References}
\renewcommand{\bibsection}{} 

\newcommand{\newblock}{} 
\bibliographystyle{dcu}
\footnotesize 
\bibliography{IEEEabrv,refs}

@STRING{IEEE_J_IP         = "{IEEE} Trans. Image Process."}

@STRING{IEEE_J_PAMI       = "{IEEE} Trans. Pattern Anal. Mach. Intell."}

@STRING{IEEE_J_BME        = "{IEEE} Trans. Biomed. Eng."}

@STRING{IEEE_J_MI         = "{IEEE} Trans. Med. Imag."}

@STRING{CLIN_ORTHO_REL_RES = "Clin. Orthop. Relat. Res."}

@STRING{ACTA_ORTHO = "Acta Orthop."}

@STRING{ACTA_RAD = "Acta Radiol."}

@STRING{ACTA_CHIR_SCAND = "Acta Chir Scand."}

@STRING{COMP_AID_SURGERY = "Comput. Aided Surg."}

@STRING{JORTHO_SURG_RES = "J. Orthop. Surg. Res."}

@STRING{IJCARS = "Int. J. Comput. Assist. Radiol. Surg."}

@STRING{IJCV = "Int. J. Comput. Vis."}

@STRING{MIA = "Med. Image Anal."}

@STRING{JBJS_AM = "J. Bone Joint Surg.-Am Vol."}

@STRING{JBJS_BR = "J. Bone Joint Surg.-Br Vol."}

@STRING{COMP_AID_SURG = "Comput. Aided Surg."}

@STRING{INTL_ORTHO = "Int. Orthop."}

@STRING{ORTH_SCI = "J. Orthop. Sci."}

@STRING{JOSA_A = "J. Opt. Soc. Am. A"}

@STRING{FRONT_BIO = "Front. Bioeng. Biotechnol."}

@STRING{MED_PHYS = "Med. Phys."}

@STRING{JBIOMECH = "J. Biomechan."}

@STRING{CAOS_BJJ = "Orthop. Proc."}

@STRING{COMM_ACM = "Commun. ACM"}

@STRING{RAD_ONC_BIO_PHYS = "Int. J. Radiat. Oncol. Biol. Phys."}

@STRING{PAT_RECOG = "Pattern Recognit."}

@STRING{PLOS_ONE = "PLoS ONE"}

@STRING{ISPRS_J_PHOTO_REMOTE_SENS = "ISPRS J. Photogramm."}

@STRING{J_ARTHRO = "J. Arthroplasty"}

@STRING{JAAOS = "J. Am. Acad. Orthop. Surg."}

@STRING{HIP_INTL = "Hip Int."}

@STRING{MICCAI = "Proc. Med. Image Comput. Comput.-Assist. Interv"}

@STRING{SPIE_MI = "Proc. {SPIE}"}

@STRING{IEEE_ISBI = "Proc. {IEEE} Intl. Symp. Biomed. Imag"}

@STRING{CVPR = "Proc. {IEEE} Conf. Comput. Vis. Pattern Recognit."}

@article{ganz1988new,
  title={A New Periacetabular Osteotomy for the Treatment of Hip Dysplasias Technique and Preliminary Results.},
  author={Ganz, Reinhold and Klaue, Kaj and Vinh, Tho Son and Mast, Jeffrey W},
  journal=CLIN_ORTHO_REL_RES,
  volume={232},
  pages={26--36},
  year={1988},
  publisher={LWW}
}

@article{gala2016hip,
  title={Hip dysplasia in the young adult},
  author={Gala, Luca and Clohisy, John C and Beaul{\'e}, Paul E},
  journal=JBJS_AM,
  volume={98},
  number={1},
  pages={63--73},
  year={2016},
  publisher={LWW}
}

@article{armand2005outcome,
  title={Outcome of periacetabular osteotomy: joint contact pressure calculation using standing {AP} radiographs, 12 patients followed for average 2 years},
  author={Armand, Mehran and Lepist{\"o}, Jyri and Tallroth, Kaj and Elias, John and Chao, Edmund},
  journal=ACTA_ORTHO,
  volume={76},
  number={3},
  pages={303--313},
  year={2005},
  publisher={Taylor \& Francis}
}

@article{armiger2009three,
  title={Three-dimensional mechanical evaluation of joint contact pressure in 12 periacetabular osteotomy patients with 10-year follow-up},
  author={Armiger, Robert S and Armand, Mehran and Tallroth, Kaj and Lepist{\"o}, Jyri and Mears, Simon C},
  journal=ACTA_ORTHO,
  volume={80},
  number={2},
  pages={155--161},
  year={2009},
  publisher={Taylor \& Francis}
}

@article{niknafs2013biomechanical,
  title={Biomechanical factors in planning of periacetabular osteotomy},
  author={Niknafs, Noushin and Murphy, Ryan J and Armiger, Robert S and Lepist{\"o}, Jyri and Armand, Mehran},
  journal=FRONT_BIO,
  volume={1},
  year={2013},
  publisher={Frontiers Media SA}
}

@article{hipp1999planning,
  title={Planning acetabular redirection osteotomies based on joint contact pressures.},
  author={Hipp, John A and Sugano, Nobuhiko and Millis, Michael B and Murphy, Stephen B},
  journal=CLIN_ORTHO_REL_RES,
  volume={364},
  pages={134--143},
  year={1999},
  publisher={LWW}
}

@article{troelsen2009surgical,
  title={Surgical advances in periacetabular osteotomy for treatment of hip dysplasia in adults},
  author={Troelsen, Anders},
  journal=ACTA_ORTHO,
  volume={80},
  number={sup332},
  pages={1--33},
  year={2009},
  publisher={Taylor \& Francis}
}

@article{troelsen2008new,
  title={A new minimally invasive transsartorial approach for periacetabular osteotomy},
  author={Troelsen, Anders and Elmengaard, B and S{\o}balle, K},
  journal=JBJS_AM,
  volume={90},
  number={3},
  pages={493--498},
  year={2008},
  publisher={LWW}
}

@article{murphy2015development,
  title={Development of a biomechanical guidance system for periacetabular osteotomy},
  author={Murphy, Ryan J and Armiger, Robert S and Lepist{\"o}, Jyri and Mears, Simon C and Taylor, Russell H and Armand, Mehran},
  journal=IJCARS,
  volume={10},
  number={4},
  pages={497--508},
  year={2015},
  publisher={Springer}
}

@article{langlotz1997first,
  title={The first twelve cases of computer assisted periacetabular osteotomy},
  author={Langlotz, Frank and Stucki, Manfred and B{\"a}chler, Richard and Scheer, Carsten and Ganz, Reinhold and Berlemann, Ulrich and Nolte, Lutz-P},
  journal=COMP_AID_SURGERY,
  volume={2},
  number={6},
  pages={317--326},
  year={1997},
  publisher={Taylor \& Francis}
}

@article{langlotz1998computer,
  title={Computer assistance for pelvic osteotomies.},
  author={Langlotz, Frank and B{\"a}chler, Richard and Berlemann, Ulrich and Nolte, Lutz-Peter and Ganz, Reinhold},
  journal=CLIN_ORTHO_REL_RES,
  volume={354},
  pages={92--102},
  year={1998},
  publisher={LWW}
}

@article{murphy2016clinical,
  title={Clinical evaluation of a biomechanical guidance system for periacetabular osteotomy},
  author={Murphy, Ryan J and Armiger, Robert S and Lepist{\"o}, Jyri and Armand, Mehran},
  journal=JORTHO_SURG_RES,
  volume={11},
  number={1},
  pages={1},
  year={2016},
  publisher={BioMed Central}
}

@article{mayman2002kingston,
  title={The Kingston periacetabular osteotomy utilizing computer enhancement: a new technique},
  author={Mayman, David J and Rudan, John and Yach, Jeff and Ellis, Randy},
  journal=COMP_AID_SURG,
  volume={7},
  number={3},
  pages={179--186},
  year={2002},
  publisher={Wiley Online Library}
}

@article{radermacher1998computer,
  title={Computer assisted orthopaedic surgery with image based individual templates.},
  author={Radermacher, Klaus and Portheine, Frank and Anton, Marc and Zimolong, Andreas and Kaspers, G{\"u}nther and Rau, G{\"u}nter and Staudte, Hans-Walter},
  journal=CLIN_ORTHO_REL_RES,
  volume={354},
  pages={28--38},
  year={1998},
  publisher={LWW}
}

@article{otsuki2013developing,
  title={Developing a novel custom cutting guide for curved peri-acetabular osteotomy},
  author={Otsuki, Bungo and Takemoto, Mitsuru and Kawanabe, Keiichi and Awa, Yasunari and Akiyama, Haruhiko and Fujibayashi, Shunsuke and Nakamura, Takashi and Matsuda, Shuichi},
  journal=INTL_ORTHO,
  volume={37},
  number={6},
  pages={1033--1038},
  year={2013},
  publisher={Springer}
}

@article{liu2016periacetabular,
  title={Periacetabular osteotomy through the pararectus approach: technical feasibility and control of fragment mobility by a validated surgical navigation system in a cadaver experiment},
  author={Liu, Li and Zheng, Guoyan and Bastian, Johannes Dominik and Keel, Marius Johann Baptist and Nolte, Lutz Peter and Siebenrock, Klaus Arno and Ecker, Timo Michael},
  journal=INTL_ORTHO,
  volume={40},
  number={7},
  pages={1389--1396},
  year={2016b},
  publisher={Springer}
}

@article{akiyama2010computed,
  title={Computed tomography-based navigation for curved periacetabular osteotomy},
  author={Akiyama, Haruhiko and Goto, Koji and So, Kazutaka and Nakamura, Takashi},
  journal=ORTH_SCI,
  volume={15},
  number={6},
  pages={829--833},
  year={2010},
  publisher={Springer}
}

@article{takao2017comparison,
  title={Comparison of rotational acetabular osteotomy performed with navigation by surgeons with different levels of experience of osteotomies},
  author={Takao, Masaki and Nishii, Takashi and Sakai, Takashi and Sugano, Nobuhiko},
  journal=IJCARS,
  volume={12},
  number={5},
  pages={841--853},
  year={2017},
  publisher={Springer}
}

@article{de2018reliability,
  title={Reliability of computer-assisted periacetabular osteotomy using a minimally invasive approach},
  author={De Raedt, Sepp and Mechlenburg, Inger and Stilling, Maiken and R{\o}mer, Lone and Murphy, Ryan J and Armand, Mehran and Lepist{\"o}, Jyri and de Bruijne, Marleen and S{\o}balle, Kjeld},
  journal=IJCARS,
  volume={13},
  number={12},
  pages={2021--2028},
  year={2018},
  publisher={Springer}
}

@article{wiberg1939studies,
  title={Studies on dysplastic acetabulum and congenital subluxation of the hip joint with special reference to the complications of osteoarthritis},
  author={Wiberg, G},
  journal=ACTA_CHIR_SCAND,
  volume={83},
  number={58},
  year={1939}
}

@article{hartig2012factors,
  title={What factors predict failure 4 to 12 years after periacetabular osteotomy?},
  author={Hartig-Andreasen, Charlotte and Troelsen, Anders and Thillemann, Theis Muncholm and S{\o}balle, Kjeld},
  journal=CLIN_ORTHO_REL_RES,
  volume={470},
  number={11},
  pages={2978--2987},
  year={2012},
  publisher={Springer}
}

@inproceedings{sadowsky2007deformable,
  title={Deformable {2D}-{3D} registration of the pelvis with a limited field of view, using shape statistics},
  author={Sadowsky, Ofri and Chintalapani, Gouthami and Taylor, Russell},
  booktitle=MICCAI,
  pages={519--526},
  year={2007}
}

@article{kang2016simultaneous,
  title={Simultaneous pose estimation and patient-specific model reconstruction from single image using maximum penalized likelihood estimation ({MPLE})},
  author={Kang, Xin and Yau, Wai-Pan and Taylor, Russell H},
  journal=PAT_RECOG,
  volume={57},
  pages={61--69},
  year={2016},
  publisher={Elsevier}
}

@article{hurvitz2008registration,
  title={Registration of a {CT}-like atlas to fluoroscopic {X}-ray images using intensity correspondences},
  author={Hurvitz, Aviv and Joskowicz, Leo},
  journal=IJCARS,
  volume={3},
  number={6},
  pages={493},
  year={2008},
  publisher={Springer}
}

@article{zheng2010statistical,
  title={Statistical shape model-based reconstruction of a scaled, patient-specific surface model of the pelvis from a single standard {AP} {X}-ray radiograph},
  author={Zheng, Guoyan},
  journal=MED_PHYS,
  volume={37},
  number={4},
  pages={1424--1439},
  year={2010},
  publisher={Wiley Online Library}
}

@article{powell2009bobyqa,
  title={The {BOBYQA} algorithm for bound constrained optimization without derivatives},
  author={Powell, Michael JD},
  journal={Cambridge NA Report NA2009/06, University of Cambridge, Cambridge},
  year={2009},
  publisher={Citeseer}
}

@inproceedings{murphy2013computer,
  title={Computer-assisted {X}-ray image-based navigation of periacetabular osteotomy with fiducial based {3D} acetabular fragment tracking},
  author={Murphy, RJ and Otake, Y and Lepist{\"o}, J and Armand, M},
  booktitle=CAOS_BJJ,
  volume={95},
  number={SUPP\_28},
  pages={84--84},
  year={2013},
  organization={The British Editorial Society of Bone \& Joint Surgery}
}

@incollection{armand2018biomechanical,
  title={Biomechanical Guidance System for Periacetabular Osteotomy},
  author={Armand, Mehran and Grupp, Robert and Murphy, Ryan and Hegman, Rachel and Armiger, Robert and Taylor, Russell and McArthur, Benjamin and Lepisto, Jyri},
  booktitle={Intelligent Orthopaedics},
  pages={169--179},
  year={2018},
  publisher={Springer}
}

@article{grupp2019pose,
  title={Pose Estimation of Periacetabular Osteotomy Fragments with Intraoperative {X}-ray Navigation},
  author={Grupp, Robert Bruce and Hegeman, Rachel and Murphy, Ryan and Alexander, Clayton and Otake, Yoshito and McArthur, Benjamin and Armand, Mehran and Taylor, Russell H},
  volume={67}, number={2}, pages={441-452},
  journal=IEEE_J_BME,
  year={2020a},
  publisher={IEEE}
}

@article{liu2016evaluation,
  title={Evaluation of constant thickness cartilage models vs. patient specific cartilage models for an optimized computer-assisted planning of periacetabular osteotomy},
  author={Liu, Li and Ecker, Timo Michael and Schumann, Steffen and Siebenrock, Klaus-Arno and Zheng, Guoyan},
  journal=PLOS_ONE,
  volume={11},
  number={1},
  pages={e0146452},
  year={2016a},
  publisher={Public Library of Science}
}

@inproceedings{krvcah2011fully,
  title={Fully automatic and fast segmentation of the femur bone from {3D}-{CT} images with no shape prior},
  author={Kr{\v{c}}ah, Marcel and Sz{\'e}kely, G{\'a}bor and Blanc, R{\'e}mi},
  booktitle=IEEE_ISBI,
  pages={2087--2090},
  year={2011}
}

@inproceedings{nikou2000description,
  title={Description of anatomic coordinate systems and rationale for use in an image-guided total hip replacement system},
  author={Nikou, Constantinos and Jaramaz, Branislav and DiGioia, Anthony M and Levison, Timothy J},
  booktitle=MICCAI,
  pages={1188--1194},
  year={2000}
}

@inproceedings{grupp2016pelvis,
  title={Pelvis surface estimation from partial {CT} for computer-aided pelvic osteotomies},
  author={Grupp, R and Otake, Y and Murphy, R and Parvizi, J and Armand, M and Taylor, R},
  booktitle=CAOS_BJJ,
  volume={98},
  number={SUPP\_5},
  pages={55--55},
  year={2016},
  organization={The British Editorial Society of Bone \& Joint Surgery}
}

@article{chintalapani2010statistical,
  title={Statistical atlas based extrapolation of {CT} data},
  author={Chintalapani, Gouthami and Murphy, Ryan and Armiger, Robert S and Lepisto, Jyri and Otake, Yoshito and Sugano, Nobuhiko and Taylor, Russell H and Armand, Mehran},
  journal=SPIE_MI,
  pages={762539--762539},
  year={2010}
}

@inproceedings{otake2015comparison,
  title={Comparison of optimization strategy and similarity metric in atlas-to-subject registration using statistical deformation model},
  author={Otake, Y and Murphy, RJ and Grupp, RB and Sato, Y and Taylor, RH and Armand, M},
  booktitle=SPIE_MI,
  pages={94150Q--94150Q},
  year={2015},
  organization={International Society for Optics and Photonics}
}

@article{horn1987closed,
  title={Closed-form solution of absolute orientation using unit quaternions},
  author={Horn, Berthold KP},
  journal=JOSA_A,
  volume={4},
  number={4},
  pages={629--642},
  year={1987},
  publisher={Optical Society of America}
}

@article{bier2019learning,
  title={Learning to detect anatomical landmarks of the pelvis in {X}-rays from arbitrary views},
  author={Bier, Bastian and Goldmann, Florian and Zaech, Jan-Nico and Fotouhi, Javad and Hegeman, Rachel and Grupp, Robert and Armand, Mehran and Osgood, Greg and Navab, Nassir and Maier, Andreas and Unberath, Mathias},
  journal=IJCARS,
  volume={14},
  number={9},
  pages={1463--1473},
  year={2019},
  publisher={Springer}
}

@article{grupp2019automatic,
  title={Automatic Annotation of Hip Anatomy in Fluoroscopy for Robust and Efficient {2D}/{3D} Registration},
  author={Grupp, Robert and Unberath, Mathias and Gao, Cong and Hegeman, Rachel and Murphy, Ryan and Alexander, Clayton and Otake, Yoshito and McArthur, Benjamin and Armand, Mehran and Taylor, Russell},
  journal=IJCARS,
  volume={15},
  number={6},
  pages={759--769},
  year={2020b}
}

@book{hartley2003multiple,
  title={Multiple view geometry in computer vision},
  author={Hartley, Richard and Zisserman, Andrew},
  year={2003},
  publisher={Cambridge university press}
}

@article{fischler1981random,
  title={Random sample consensus: a paradigm for model fitting with applications to image analysis and automated cartography},
  author={Fischler, Martin A and Bolles, Robert C},
  journal=COMM_ACM,
  volume={24},
  number={6},
  pages={381--395},
  year={1981},
  publisher={ACM}
}

@article{loy2003fast,
  title={Fast radial symmetry for detecting points of interest},
  author={Loy, Gareth and Zelinsky, Alexander},
  journal=IEEE_J_PAMI,
  volume={25},
  number={8},
  pages={959--973},
  year={2003},
  publisher={IEEE}
}

@article{blondel2006reconstruction,
  title={Reconstruction of coronary arteries from a single rotational {X}-ray projection sequence},
  author={Blondel, Christophe and Malandain, Gr{\'e}goire and Vaillant, R{\'e}gis and Ayache, Nicholas},
  journal=IEEE_J_MI,
  volume={25},
  number={5},
  pages={653--663},
  year={2006},
  publisher={IEEE}
}

@article{yang2009novel,
  title={Novel approach for 3-D reconstruction of coronary arteries from two uncalibrated angiographic images},
  author={Yang, Jian and Wang, Yongtian and Liu, Yue and Tang, Songyuan and Chen, Wufan},
  journal=IEEE_J_IP,
  volume={18},
  number={7},
  pages={1563--1572},
  year={2009},
  publisher={IEEE}
}

@article{dang2012robust,
  title={Robust methods for automatic image-to-world registration in cone-beam {CT} interventional guidance},
  author={Dang, H and Otake, Yo and Schafer, S and Stayman, Joseph Webster and Kleinszig, G and Siewerdsen, JH},
  journal=MED_PHYS,
  volume={39},
  number={10},
  pages={6484--6498},
  year={2012},
  publisher={Wiley Online Library}
}

@article{hamming2009automatic,
  title={Automatic image-to-world registration based on {X}-ray projections in cone-beam {CT}-guided interventions},
  author={Hamming, NM and Daly, MJ and Irish, JC and Siewerdsen, JH},
  journal=MED_PHYS,
  volume={36},
  number={5},
  pages={1800--1812},
  year={2009},
  publisher={Wiley Online Library}
}

@article{choi2014fiducial,
  title={Fiducial marker-based correction for involuntary motion in weight-bearing {C}-arm {CT} scanning of knees. {II}. {E}xperiment},
  author={Choi, Jang-Hwan and Maier, Andreas and Keil, Andreas and Pal, Saikat and McWalter, Emily J and Beaupr{\'e}, Gary S and Gold, Garry E and Fahrig, Rebecca},
  journal=MED_PHYS,
  volume={41},
  number={6Part1},
  pages={061902},
  year={2014},
  publisher={Wiley Online Library}
}

@article{yaniv2009localizing,
  title={Localizing spherical fiducials in {C}-arm based cone-beam {CT}},
  author={Yaniv, Ziv},
  journal=MED_PHYS,
  volume={36},
  number={11},
  pages={4957--4966},
  year={2009},
  publisher={Wiley Online Library}
}

@article{selvik1990roentgen,
  title={Roentgen stereophotogrammetric analysis},
  author={Selvik, G{\"o}ran},
  journal=ACTA_RAD,
  volume={31},
  number={2},
  pages={113--126},
  year={1990},
  publisher={Taylor \& Francis}
}

@article{valstar2002use,
  title={The use of Roentgen stereophotogrammetry to study micromotion of orthopaedic implants},
  author={Valstar, Edward R and Nelissen, Rob GHH and Reiber, Johan HC and Rozing, Piet M},
  journal=ISPRS_J_PHOTO_REMOTE_SENS,
  volume={56},
  number={5-6},
  pages={376--389},
  year={2002},
  publisher={Elsevier}
}

@article{karrholm1984changes,
  title={Changes in tibiofibular relationships due to growth disturbances after ankle fractures in children.},
  author={K{\"a}rrholm, J and Hansson, LARS INGVAR and Selvik, G},
  journal=JBJS_AM,
  volume={66},
  number={8},
  pages={1198--1210},
  year={1984}
}

@article{de2008image,
  title={Image-based {RSA}: Roentgen stereophotogrammetric analysis based on {2D}--{3D} image registration},
  author={De Bruin, PW and Kaptein, BL and Stoel, BC and Reiber, JHC and Rozing, PM and Valstar, ER},
  journal=JBIOMECH,
  volume={41},
  number={1},
  pages={155--164},
  year={2008},
  publisher={Elsevier}
}

@article{seehaus2012markerless,
  title={Markerless Roentgen Stereophotogrammetric Analysis for in vivo implant migration measurement using three dimensional surface models to represent bone},
  author={Seehaus, Frank and Olender, Gavin D and Kaptein, Bart L and Ostermeier, Sven and Hurschler, Christof},
  journal=JBIOMECH,
  volume={45},
  number={8},
  pages={1540--1545},
  year={2012},
  publisher={Elsevier}
}

@article{mechlenburg2007safe,
  title={Safe fixation with two acetabular screws after {G}anz periacetabular osteotomy},
  author={Mechlenburg, Inger and Kold, S{\o}ren and R{\o}mer, Lone and S{\o}balle, Kjeld},
  journal=ACTA_ORTHO,
  volume={78},
  number={3},
  pages={344--349},
  year={2007},
  publisher={Taylor \& Francis}
}

@article{yuan2002roentgen,
  title={Roentgen single-plane photogrammetric analysis ({RSPA}) a new approach to the study of musculoskeletal movement},
  author={Yuan, X and Ryd, Leif and Tanner, KE and Lidgren, Lars},
  journal=JBJS_BR,
  volume={84},
  number={6},
  pages={908--914},
  year={2002},
  publisher={The British Editorial Society of Bone and Joint Surgery}
}

@inproceedings{steger2013marker,
  title={Marker detection evaluation by phantom and cadaver experiments for {C}-arm pose estimation pattern},
  author={Steger, Teena and Ho{\ss}bach, Martin and Wesarg, Stefan},
  booktitle=SPIE_MI,
  volume={8671},
  pages={86711V},
  year={2013},
  organization={International Society for Optics and Photonics}
}

@article{jain2005ftrac,
  title={{FTRAC}--{A} robust fluoroscope tracking fiducial},
  author={Jain, Ameet Kumar and Mustafa, Tabish and Zhou, Yu and Burdette, Clif and Chirikjian, Gregory S and Fichtinger, Gabor},
  journal=MED_PHYS,
  volume={32},
  number={10},
  pages={3185--3198},
  year={2005a},
  publisher={Wiley Online Library}
}

@article{kang2013robustness,
  title={Robustness and accuracy of feature-based single image 2-{D}--3-{D} registration without correspondences for image-guided intervention},
  author={Kang, Xin and Armand, Mehran and Otake, Yoshito and Yau, Wai-Pan and Cheung, Paul YS and Hu, Yong and Taylor, Russell H},
  journal=IEEE_J_BME,
  volume={61},
  number={1},
  pages={149--161},
  year={2013},
  publisher={IEEE}
}

@article{aubry2004measurements,
  title={Measurements of intrafraction motion and interfraction and intrafraction rotation of prostate by three-dimensional analysis of daily portal imaging with radiopaque markers},
  author={Aubry, Jean-Francois and Beaulieu, Luc and Girouard, Louis-Martin and Aubin, Sylviane and Tremblay, Daniel and Laverdi{\`e}re, Jacques and Vigneault, Eric},
  journal=RAD_ONC_BIO_PHYS,
  volume={60},
  number={1},
  pages={30--39},
  year={2004},
  publisher={Elsevier}
}

@article{litzenberg2002daily,
  title={Daily prostate targeting using implanted radiopaque markers},
  author={Litzenberg, Dale and Dawson, Laura A and Sandler, Howard and Sanda, Martin G and McShan, Daniel L and Ten Haken, Randall K and Lam, Kwok L and Brock, Kristy K and Balter, James M},
  journal=RAD_ONC_BIO_PHYS,
  volume={52},
  number={3},
  pages={699--703},
  year={2002},
  publisher={Elsevier}
}

@article{schweikard2000robotic,
  title={Robotic motion compensation for respiratory movement during radiosurgery},
  author={Schweikard, Achim and Glosser, Greg and Bodduluri, Mohan and Murphy, Martin J and Adler, John R},
  journal=COMP_AID_SURGERY,
  volume={5},
  number={4},
  pages={263--277},
  year={2000},
  publisher={Wiley Online Library}
}

@inproceedings{tang2000fiducial,
  title={Fiducial registration from a single X-Ray image: a new technique for fluoroscopic guidance and radiotherapy},
  author={Tang, Thomas SY and Ellis, Randy E and Fichtinger, Gabor},
  booktitle=MICCAI,
  pages={502--511},
  year={2000},
  organization={Springer}
}

@article{tang2004accurate,
  title={Accurate assessment of patellar tracking using fiducial and intensity-based fluoroscopic techniques},
  author={Tang, Thomas SY and MacIntyre, NJ and Gill, HS and Fellows, RA and Hill, NA and Wilson, DR and Ellis, Randy E},
  journal=MIA,
  volume={8},
  number={3},
  pages={343--351},
  year={2004},
  publisher={Elsevier}
}

@article{ioppolo2007validation,
  title={Validation of a low-dose hybrid {RSA} and fluoroscopy technique: {D}etermination of accuracy, bias and precision},
  author={Ioppolo, James and B{\"o}rlin, Niclas and Bragdon, Charles and Li, Mingguo and Price, Roger and Wood, David and Malchau, Henrik and Nivbrant, Bo},
  journal=JBIOMECH,
  volume={40},
  number={3},
  pages={686--692},
  year={2007},
  publisher={Elsevier}
}

@article{snavely2008modeling,
  title={Modeling the world from internet photo collections},
  author={Snavely, Noah and Seitz, Steven M and Szeliski, Richard},
  journal=IJCV,
  volume={80},
  number={2},
  pages={189--210},
  year={2008},
  publisher={Springer}
}

@article{westoby2012structure,
  title={`{S}tructure-from-Motion' photogrammetry: A low-cost, effective tool for geoscience applications},
  author={Westoby, Matthew J and Brasington, James and Glasser, Niel F and Hambrey, Michael J and Reynolds, JM},
  journal={Geomorphology},
  volume={179},
  pages={300--314},
  year={2012},
  publisher={Elsevier}
}

@inproceedings{schonberger2016structure,
  title={Structure-from-motion revisited},
  author={Schonberger, Johannes L and Frahm, Jan-Michael},
  booktitle=CVPR,
  pages={4104--4113},
  year={2016}
}

@article{mechlenburg2009radiation,
  title={Radiation exposure to the orthopaedic surgeon during periacetabular osteotomy},
  author={Mechlenburg, Inger and Daugaard, Henrik and S{\o}balle, Kjeld},
  journal=INTL_ORTHO,
  volume={33},
  number={6},
  pages={1747},
  year={2009},
  publisher={Springer}
}

@article{west2001fiducial,
  title={Fiducial point placement and the accuracy of point-based, rigid body registration},
  author={West, Jay B and Fitzpatrick, J Michael and Toms, Steven A and Maurer Jr, Calvin R and Maciunas, Robert J},
  journal={Neurosurgery},
  volume={48},
  number={4},
  pages={810--817},
  year={2001},
  publisher={Oxford University Press}
}

@inproceedings{jain2005c,
  title={C-arm calibration--Is it really necessary?},
  author={Jain, Ameet and Kon, Ryan and Zhou, Yu and Fichtinger, Gabor},
  booktitle=MICCAI,
  pages={639--646},
  year={2005b},
  organization={Springer}
}

@article{stallmann2005biodegradable,
  title={Biodegradable {X}-ray markers of controlled radio-opacity: Temporary position measurements in bone},
  author={Stallmann, Hein P and Faber, Chris and Plokker, Herbert M and Wuisman, Paul IJM},
  journal=ACTA_ORTHO,
  volume={76},
  number={1},
  pages={122--127},
  year={2005},
  publisher={Taylor \& Francis}
}

@article{lawrie2003insertion,
  title={Insertion of tantalum beads in {RSA} of the hip Variations in incidence of extra-osseous beads with insertion site},
  author={Lawrie, David and Downing, Martin and Ashcroft, G Patrick and Gibson, Peter},
  journal=ACTA_ORTHO,
  volume={74},
  number={4},
  pages={404--407},
  year={2003},
  publisher={Taylor \& Francis}
}

@article{shah2018routine,
  title={Routine use of Radiostereometric analysis in elective hip and knee arthroplasty patients: surgical impact, safety, and bead stability},
  author={Shah, Roshan P and MacLean, Leanna and Paprosky, Wayne G and Sporer, Scott},
  journal=JAAOS,
  volume={26},
  number={8},
  pages={e173--e180},
  year={2018},
  publisher={LWW}
}

@article{nieuwenhuijse2012good,
  title={Good diagnostic performance of early migration as a predictor of late aseptic loosening of acetabular cups: results from ten years of follow-up with {R}oentgen stereophotogrammetric analysis ({RSA})},
  author={Nieuwenhuijse, Marc J and Valstar, Edward R and Kaptein, Bart L and Nelissen, Rob GHH},
  journal=JBJS_AM,
  volume={94},
  number={10},
  pages={874--880},
  year={2012},
  publisher={LWW}
}

@article{klerken2015high,
  title={High early migration of the revised acetabular component is a predictor of late cup loosening: 312 cup revisions followed with radiostereometric analysis for 2-20 years},
  author={Klerken, Tina and Mohaddes, Maziar and Nemes, Szilard and K{\"a}rrholm, Johan},
  journal=HIP_INTL,
  volume={25},
  number={5},
  pages={471--476},
  year={2015},
  publisher={SAGE Publications Sage UK: London, England}
}

@article{otten2016stability,
  title={Stability of uncemented cupsÑlong-term effect of screws, pegs and {HA} coating: a 14-year {RSA} follow-up of total hip arthroplasty},
  author={Otten, Volker TC and Crnalic, Sead and R{\"o}hrl, Stephan M and Nivbrant, Bo and Nilsson, Kjell G},
  journal=J_ARTHRO,
  volume={31},
  number={1},
  pages={156--161},
  year={2016},
  publisher={Elsevier}
}
\normalsize
\renewcommand{\thesection}{S-\arabic{section}}
\renewcommand{\thetable}{S-\arabic{table}}
\renewcommand{\thefigure}{S-\arabic{figure}}
\setcounter{section}{0}
\setcounter{figure}{0}
\setcounter{table}{0}
\section{Supplementary Methods}\label{sec:supp_methods}
\subsection{BB Detection Using Fast Radial Symmetry}\label{sec:supp_methods_rad_symmetry}
BB locations are estimated in each 2D fluoroscopic image using a variant of the fast radial symmetry algorithm~\cite{loy2003fast}.
For the cases of specimen~1, with $1.5$ mm BBs, a single radius of 4 pixels is used.
For specimens~2 and~3, with 1~mm BBs, radii of~1 and 2~pixels were used.
Standard deviations of the Gaussian kernels were set to $n/4$, where $n$ is the current radius parameter.
The radial strictness parameter, $\alpha$, was set to 1.
Unlike the reference description, no normalization adjustments were made to the orientation projection and magnitude projection images in our implementation -- the $k_n$ parameter is not used.

Locations of each BB are estimated using the radial symmetry output image, $S$. 
Let $M$ denote the maximum value of $S$, and let $r'$ denote the smallest radius considered (either 4 or 1 depending on the BB size).
A pixel index, $(x,y)$, is classified as the center of a BB if $S(x,y)$ is the equal to the maximum value of $S$ in a $(2r' + 1) \times (2r' + 1)$ neighborhood about $(x,y)$, and $S(x,y) > 0.2M$.
\subsection{Intensity-Based 2D/3D Registration Parameters}\label{sec:app_intensity_regi}
The pelvis-as-fiducial, intensity-based, registration parameters described in~\cite{grupp2019pose} are exactly those used for the pre-osteotomy BB reconstruction phase.
Each registration in the reconstruction phase runs two resolutions, $8\times$ and $4\times$ downsampling in 2D.
In order to overcome large initialization offsets from ground truth, a computationally expensive, evolutionary optimization strategy is used at the $8\times$ level.
The less computation-intensive, BOBYQA strategy~\cite{powell2009bobyqa}, is used for optimization at the second level.

In order to avoid delays in the surgical workflow, a small execution time is desired during the post-osteotomy pose estimation phase.
Therefore, the BOBYQA strategy is used at a single resolution level of $8\times$ downsampling in 2D.
We believe a local optimization strategy is sufficient, since solutions reported by the P3P solver, using correct BB correspondences, should lie within some convex ball of the ground truth pose.
Box constraints on the $\mathfrak{se}(3)$ parameter space are specified as in \eqref{eq:app_bobyqa_bounds}, where the X and Y axes are roughly aligned with image columns and rows, respectively, and the Z axis is aligned with the source-to-detector axis.
\begin{equation} \label{eq:app_bobyqa_bounds}
	\left \{ \pm 15\degree, \pm 15\degree, \pm 30\degree, \pm 50, \pm 50, \pm 100 \right \}
\end{equation}
All other registration parameters remain identical to those of the reconstruction phase.
\subsection{Intraoperative BB Reconstruction}\label{sec:app_bb_recon}
\begin{figure}
\begin{indented}
\item[]
\begin{center}
\includegraphics[width=0.926\linewidth]{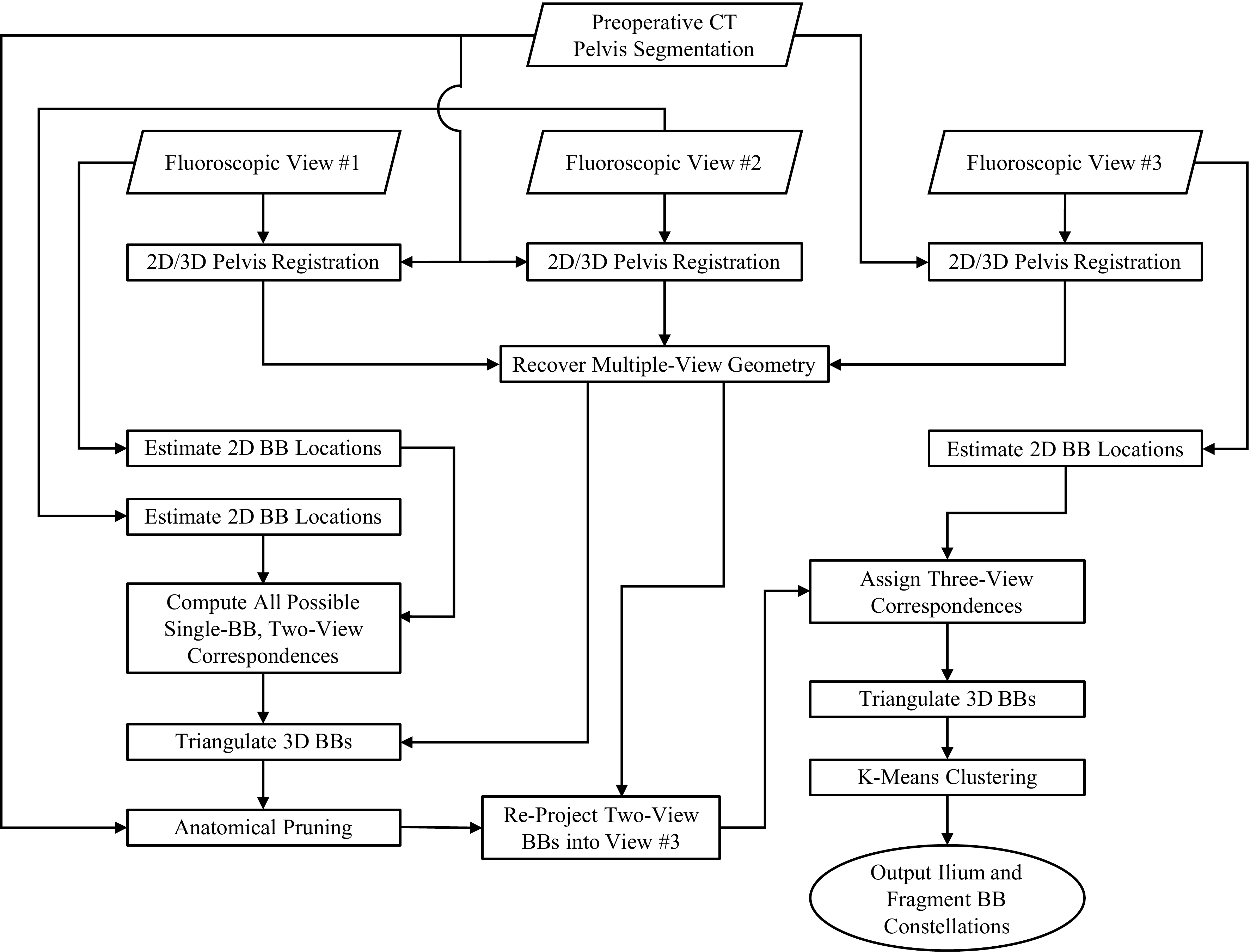}
\end{center}
\end{indented}
\caption{A workflow overview of the intraoperative BB reconstruction process.
Three separate 2D/3D pelvis registrations of each fluoroscopic view are performed to recover the relative poses of the C-arm.
Triangulations from all possible single-BB correspondences in the first two views are computed, and pruned using the 3D pelvis segmentation.
Any remaining, invalid, correspondences are eliminated by re-projecting into the third view and checking for consistency with 2D BB detections.
Using the correct three-view correspondences, the BBs are re-triangulated, and K-Means is used to label each BB as belonging to the ilium or fragment constellation.}
\label{fig:bb_recon_workflow}
\end{figure}

The workflow of the entire BB reconstruction process is shown in \fig~\ref{fig:bb_recon_workflow}.

Once the 2D BB locations have been detected in each view and the multiple-view geometry is recovered using the 2D/3D pelvis single-view registrations, the following algorithm is used to reconstruct the 3D positions of each BB.
Denote the sets of 2D detected BB locations as $P_v \subset \mathbb{R}^2$ for each view $v=1,2,3$.
Let $\mathcal{T} : \wp(\mathbb{R}^2) \rightarrow \mathbb{R}^3$ denote the triangulation operator used to reconstruct a 3D point from a collection of 2D points; $\wp$ indicates the power set operator.
Let $\mathcal{D} : \mathbb{R}^3 \rightarrow \mathbb{R}$ denote the minimum distance between a 3D point and the pelvis surface.
Let $\mathcal{P}_v : \mathbb{R}^3 \rightarrow \mathbb{R}^2$ denote the projection operator, applying a perspective projection of 3D points into the imaging plane of view $v$.
As shown in \eqref{eq:corr_init}, an initial set of correspondences and 3D triangulations, $A$, are computed for each of the candidate correspondences and any points lying further than $T$ mm away from the pelvis surface are pruned.
\begin{equation} \label{eq:corr_init}
	A = \left \{ \left(p,q, \mathcal{T} \left(p,q\right) \right) | p \in P_1, q \in P_2, \mathcal{D}\left(\mathcal{T}\left(p,q\right)\right) < T \right \}
\end{equation}
The remaining triangulated points are re-projected into the third view and the 2D distances to each BB detection are recorded, shown in \eqref{eq:corr_reproj}.
\begin{equation} \label{eq:corr_reproj}
	B = \left \{ \left( p, q, r, d \right) | (p, q, x) \in A, r \in P_3, d = \left \| \mathcal{P}_3 \left( x \right) - r \right \|_2 \right \}
\end{equation}
The following book-keeping sets are initialized: the final set of 3D reconstructed points $C = \{ \}$, and sets indicating whether a 2D BB detection has been used for 3D reconstruction, $R_v = \{ \}$ for $v = 1,2,3$.
We now iterate through $B$ in \textit{increasing} order according to the re-projection component, $d$.
For each $\left( p,q,r,d \right) \in B$, if $p \notin R_1$, $q \notin R_2$, and $r \notin R_3$, let $y = \mathcal{T} \left( p, q, r \right)$.
If $\mathcal{D}\left( y \right) < T$, then the point is a suitable reconstruction; update the book-keeping sets: $C = C \cup \{ y \}$, $R_1 = R_1 \cup \{ p \}$, $R_2 = R_2 \cup \{ q \}$, and $R_3 = R_3 \cup \{ r \}$.
Once iteration over $B$ is complete $C$ represents the final set of 3D BB reconstructions.
Iteration may be terminated early when any $R_v$ is equal to $P_v$ for $v=1,2,3$.
\subsection{Intraoperative Pose Estimation}\label{sec:app_p3p}
\begin{figure}
\begin{indented}
\item[]
\begin{center}
\includegraphics[width=\linewidth]{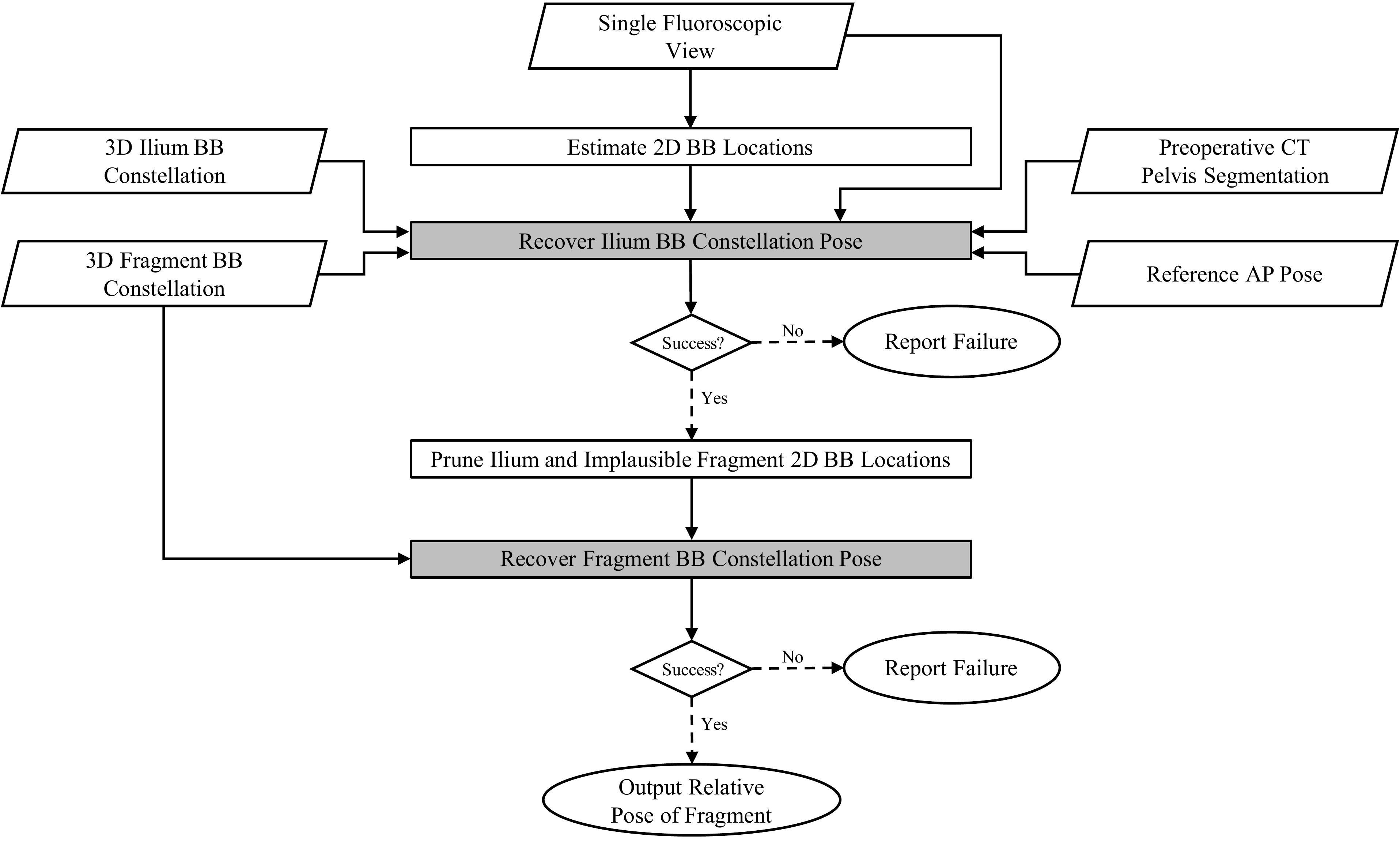}
\end{center}
\end{indented}
\caption{Complete workflow used for single-view relative pose estimation of the acetabular fragment.
Gray-shaded boxes correspond to the ilium and fragment BB constellation pose estimate workflows described in figures~\ref{fig:pose_est_workflow_ilium} and~\ref{fig:pose_est_workflow_frag}, respectively.
The relative pose of the bone fragment is calculated using the BB constellation poses.}
\label{fig:pose_est_workflow_combined}
\end{figure}

\begin{figure}
\begin{indented}
\item[]
\begin{center}
\includegraphics[width=0.926\linewidth]{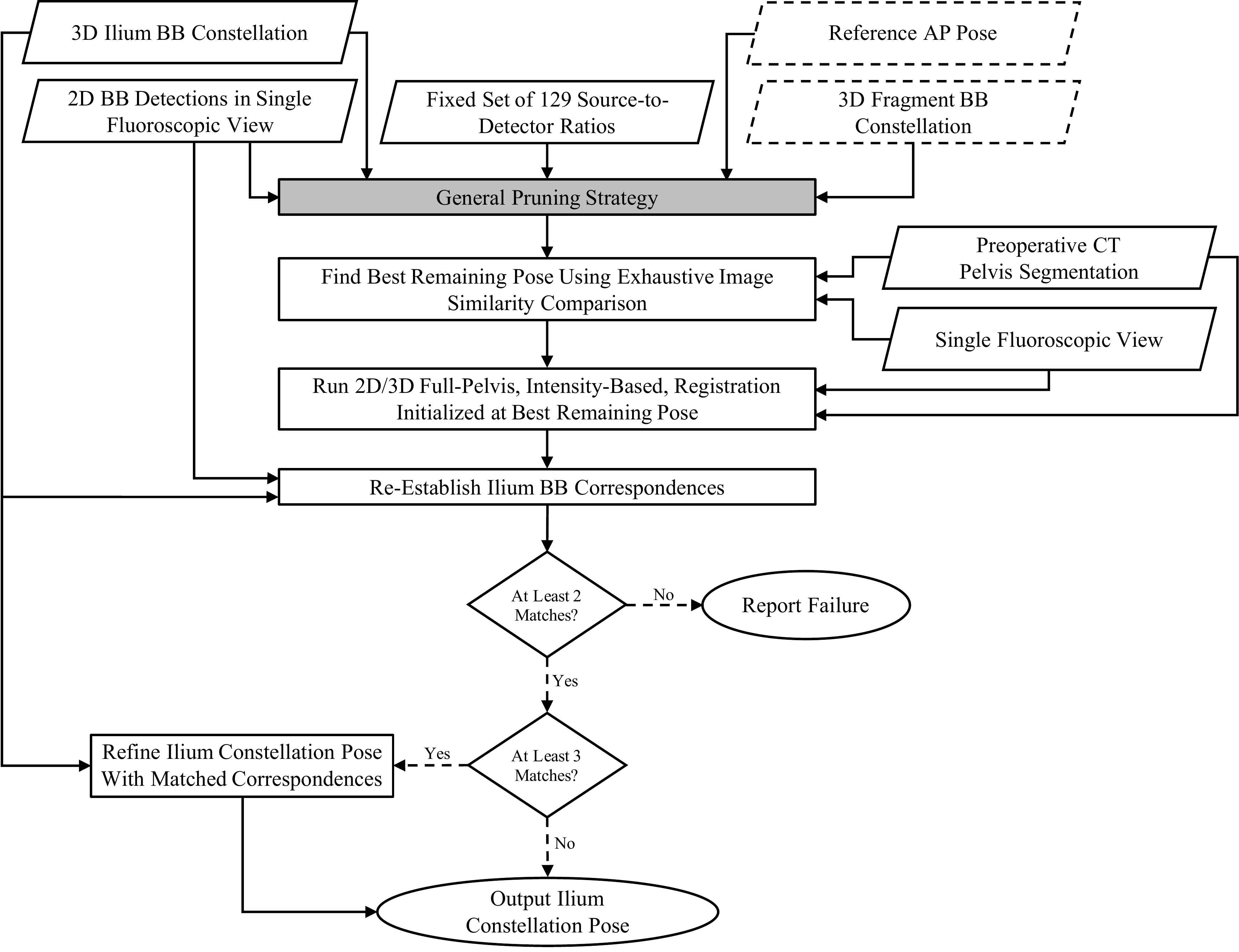}
\end{center}
\end{indented}
\caption{Overview of the ilium BB constellation pose estimation process.
The workflow of the general pruning strategy (\fig~\ref{fig:pose_est_workflow_general_pruning}) is re-used here and highlighted in gray, with inputs specific to ilium pruning emphasized by dashed borders.
Since the general pruning strategy returns multiple possible poses and BB correspondences, image intensities are used to select the best candidate pose.
The pose is further refined by a 2D/3D intensity-based registration of the pre-osteotomy pelvis, with success criteria automatically verified by the number of ilium BBs matched through re-projection.}
\label{fig:pose_est_workflow_ilium}
\end{figure}

\begin{figure}
\begin{indented}
\item[]
\begin{center}
\includegraphics[width=0.792\linewidth]{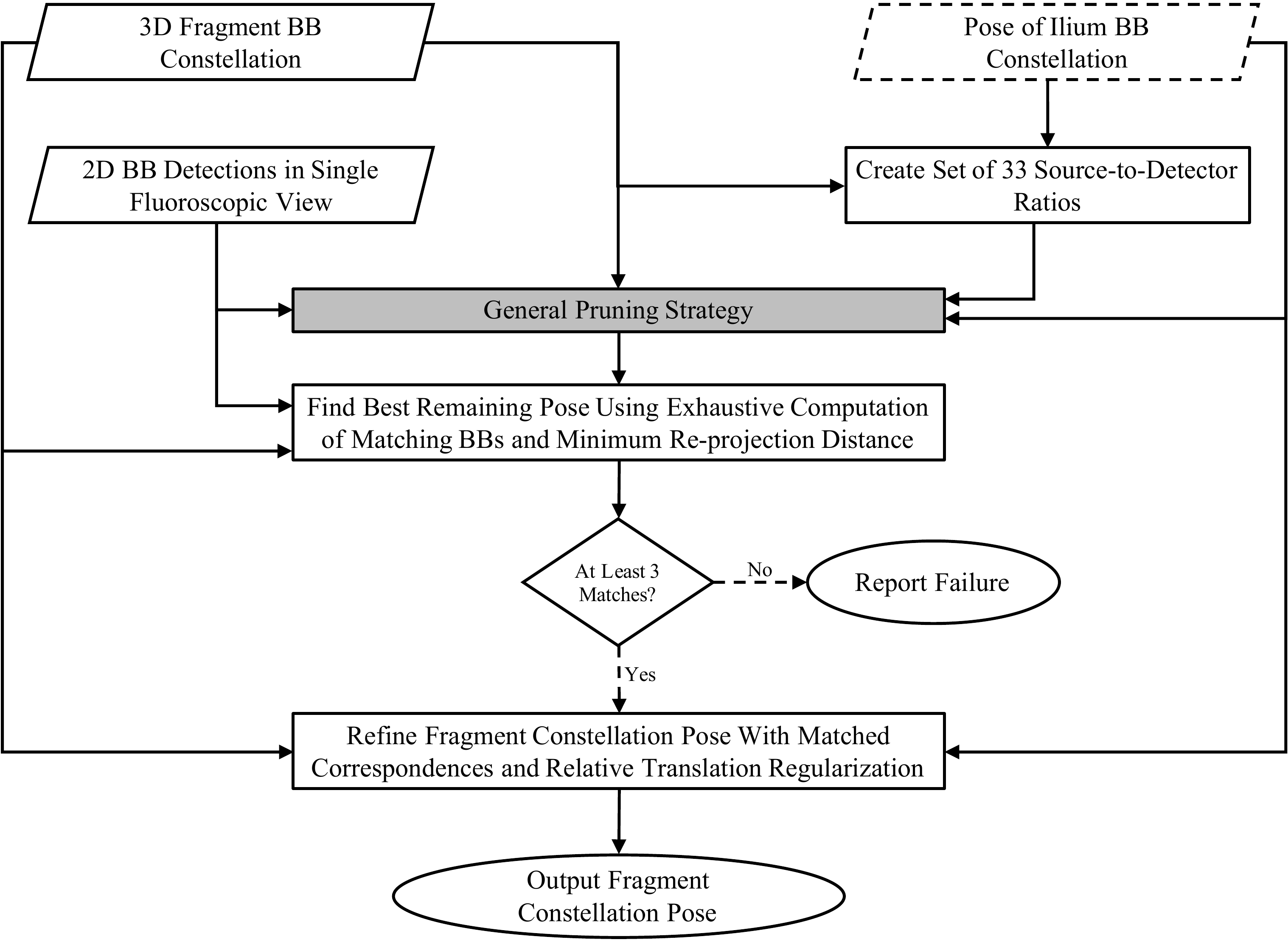}
\end{center}
\end{indented}
\caption{Workflow of the fragment BB constellation pose estimation process.
Gray shading corresponds to the invocation of the general pruning strategy (\fig~\ref{fig:pose_est_workflow_general_pruning}), with the ilium pose used to prune implausible relative fragment poses.
The best pose returned by the general strategy is selected by maximizing the number of matched re-projected fragment BBs with smallest mean in-plane, re-projection, distance.
The final pose is only reported when at least three fragment BBs are matched.}
\label{fig:pose_est_workflow_frag}
\end{figure}

\begin{figure}
\begin{indented}
\item[]
\begin{center}
\includegraphics[width=0.537\linewidth]{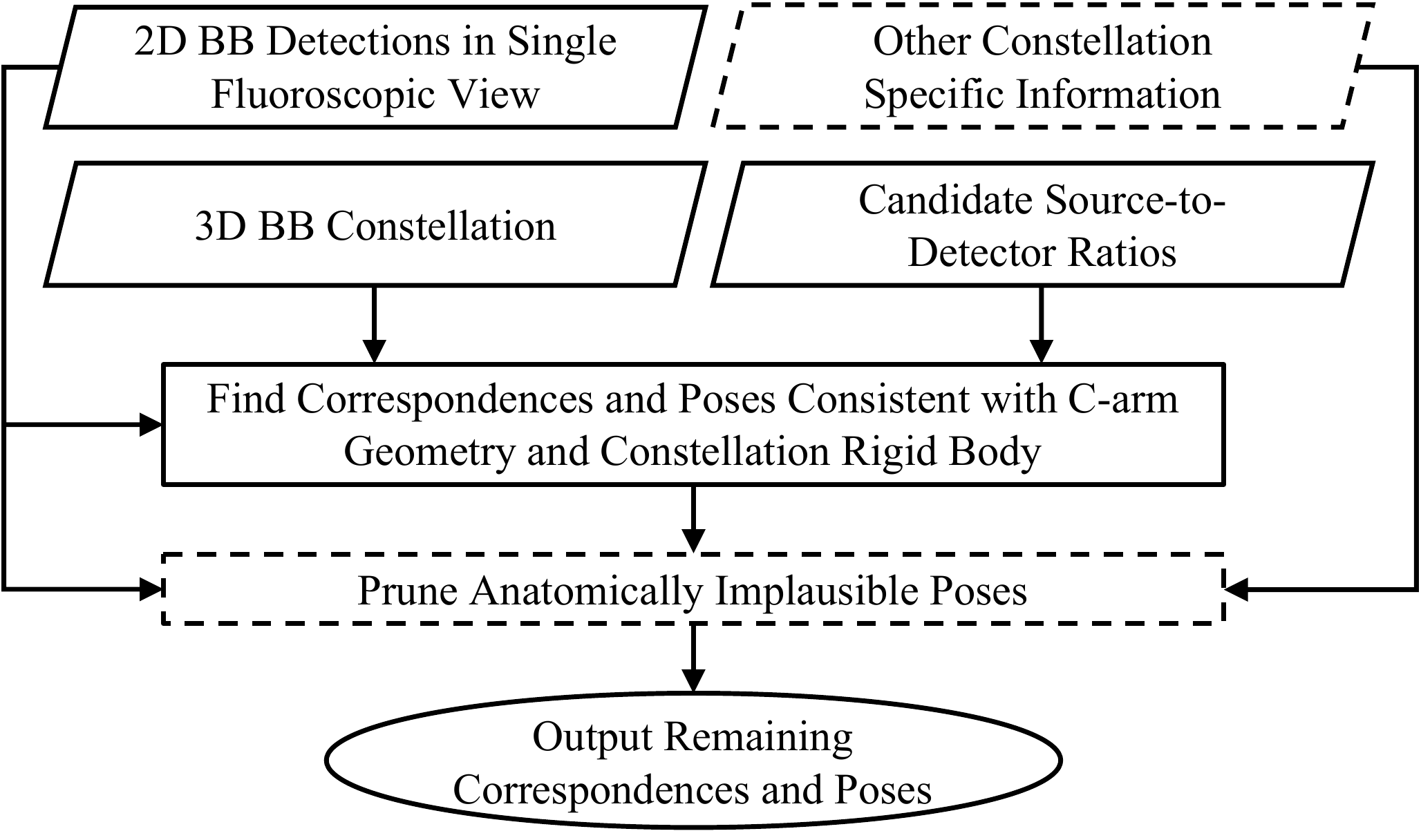}
\end{center}
\end{indented}
\caption{The data flow of the general pruning strategy used during BB constellation pose estimations.
Dashed boxes indicate input data and processing that will be specific for either ilium or fragment processing.}
\label{fig:pose_est_workflow_general_pruning}
\end{figure}

\Fig~\ref{fig:pose_est_workflow_combined} depicts the entire, end-to-end, pose estimation workflow and figures~\ref{fig:pose_est_workflow_ilium} and~\ref{fig:pose_est_workflow_frag} show the specific workflows for obtaining pose estimates of the ilium and fragment BB constellations, respectively.
\Fig~\ref{fig:pose_est_workflow_general_pruning} illustrates the general pruning strategy shared by the ilium and fragment pose estimation methods.

The general pruning strategy relies on an efficient P3P solver, which finds plausible transformations that rigidly map three 3D BB points into a C-arm coordinate frame, so that their projected locations in 2D match a set of corresponding 2D points.
Implausible poses are eliminated using constraints derived from the approximate C-arm geometry and rigid structure of the BB constellations.
The remainder of this section provides details of the P3P algorithm.

Let $B_1,B_2,B_3$ denote the 3D model points and let $b_1,b_2,b_3$ be their (hypothesized) corresponding 2D points in the fluoroscopic image.
The problem is simplified by assuming that the approximate depth, or proportion along the source-to-detector line, of $B_1$ in the C-arm frame is known.
Given this information, we know the location of $B_1$ with respect to the C-arm, denoted as $\tilde{B}_1$.
For $j = 2,3$, let $\tilde{B}_j(t) = s + t (\hat{b}_j - s)$ denote the lines which $B_2$ and $B_3$, with respect to the C-arm, may possibly lie on.
The X-ray source position is denoted by $s$ and the 3D location on the X-ray detector, corresponding to $b_j$, is denoted by $\hat{b}_j$.
For specific values of $t_2$ and $t_3$, a potential pose is given by solving the 3D/3D corresponding point set registration~\cite{horn1987closed} between $\{ \tilde{B}_1, \tilde{B}_2(t_2), \tilde{B}_3(t_3) \}$ and $\{ B_1, B_2, B_3 \}$.
We find the four possible combinations of $t_2$ and $t_3$ and use the known shape of the 3D model to prune implausible poses.

Let $l_{ij} = \| B_i - B_j \|_2$ denote an inter-BB distance of the 3D model; the $l_{ij}$ are known quantities.
Let $\tilde{l}_{1j}(t) = \|\tilde{B}_1 - \tilde{B}_j(t)\|_2$; the $\tilde{l}_{1j}(t)$ are unknown quantities.
Using $l_{12}$ and $l_{13}$, we find plausible values of $t$ for $\tilde{B}_2$ and $\tilde{B}_3$ by solving \eqref{eq:app_p3p_opt_prob} for $j=2,3$.
\begin{equation}\label{eq:app_p3p_opt_prob}
	\min_t \left( l_{1j}^2 - \tilde{l}_{1j}^2 \left( t \right) \right)^2
\end{equation}
Using MATLAB 2019a, derivatives and formulas for the possible minimizers of \eqref{eq:app_p3p_opt_prob} were symbolically calculated; a maximum of 2 minimizers are possible.
Let $t_j^{(1)}$ and $t_j^{(2)}$ denote the two solutions of \eqref{eq:app_p3p_opt_prob} for $j=2,3$.
Poses are pruned when \eqref{eq:app_p3p_t_thresh}, \eqref{eq:app_p3p_l1j_thresh} and \eqref{eq:app_p3p_lij_thresh} are not satisfied for the combinations of $j=2,3$, $k=1,2$, and $m=1,2$.
\begin{equation} \label{eq:app_p3p_t_thresh}
	0.6 \leq t_j^{(k)} \leq 1.0
\end{equation}
\begin{equation} \label{eq:app_p3p_l1j_thresh}
	1 - \epsilon \leq \frac{\tilde{l}_{1j}\left(t_j^{(k)}\right)}{l_{1j}} \leq 1 + \epsilon
\end{equation}
\begin{equation} \label{eq:app_p3p_lij_thresh}
	1 - \epsilon \leq \frac{\left \| \tilde{B}_{2}\left(t_2^{(k)}\right) - \tilde{B}_{3}\left(t_3^{(m)}\right) \right \|_2}{l_{23}} \leq 1 + \epsilon
\end{equation}
Pruning using \eqref{eq:app_p3p_t_thresh} constrains objects to lie closer to the X-ray detector than the X-ray source.
Pruning using \eqref{eq:app_p3p_l1j_thresh} and \eqref{eq:app_p3p_lij_thresh} constrains $\{ \tilde{B}_1, \tilde{B}_2(t_2), \tilde{B}_3(t_3) \}$ to have the same shape as $\{ B_1, B_2, B_3 \}$.
The pruning should be conducted in a greedy fashion in order to avoid unnecessary computation.
A toy example depicting the geometries described here is shown in \fig~\ref{fig:bb_pose_toy}.
For all experiments in this paper, $\epsilon = 0.01$.

Similar to the method of Appendix A.3 in~\cite{fischler1981random}, we perform this process over a range of source-to-detector distances in order to achieve robustness in depth for $\tilde{B}_1$.
A minimum of zero, and a maximum of four, poses are identified for each depth.
\begin{figure}
\begin{indented}
\item[]
\begin{center}
\includegraphics[width=0.75\linewidth]{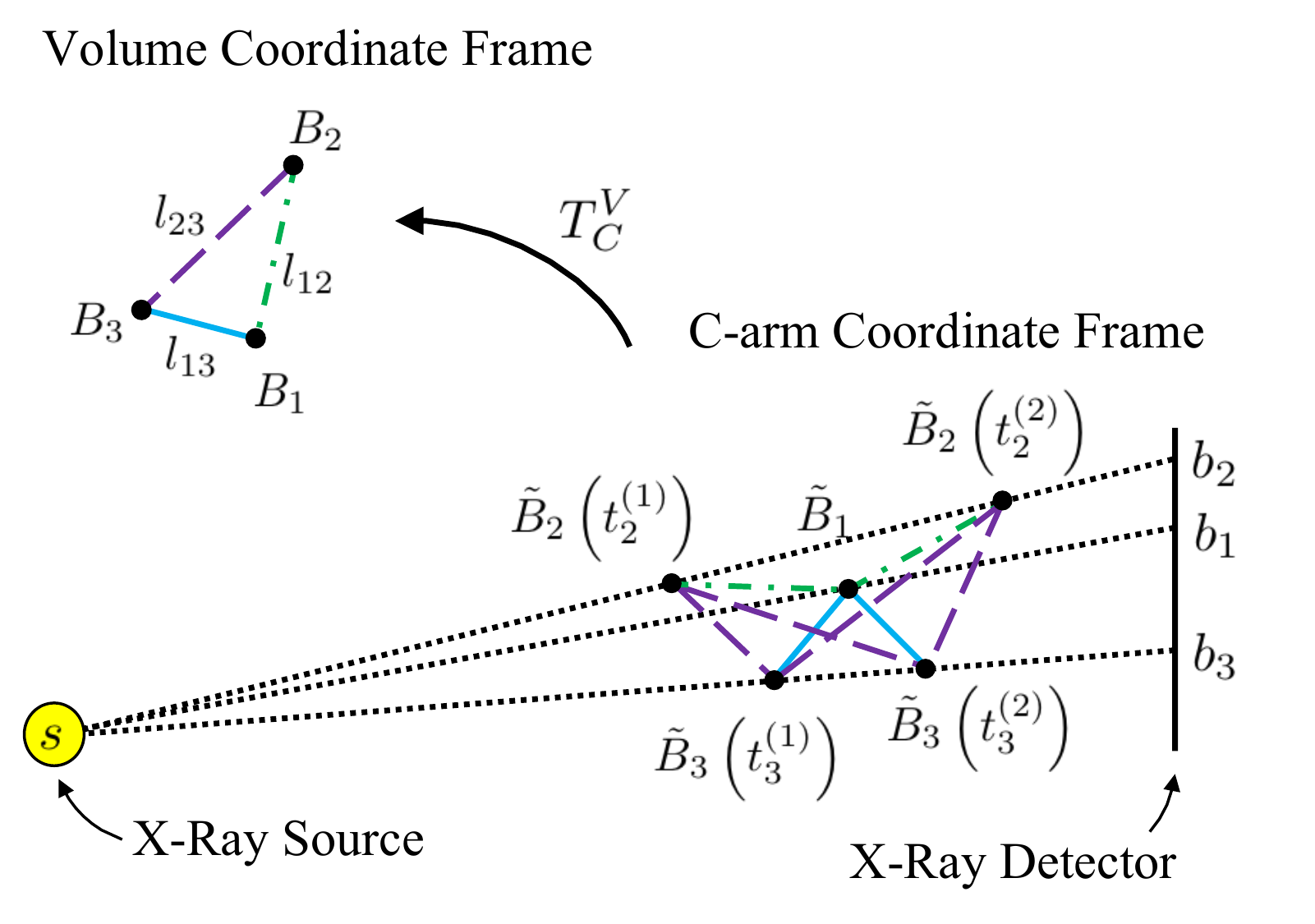}
\end{center}
\end{indented}
\caption{A toy example of the P3P problem showing the four possible solutions when mapping the BB constellation into the C-arm coordinate frame.
See the text of section~\ref{sec:app_p3p} for a full explanation of notation.
This drawing represents a specific source-to-detector distance used to estimate $\tilde{B}_1$.
For each of $B_2$ and $B_3$, two possible locations with respect to the C-arm are shown.
The inter-BB length to $B_1$ is preserved for all 4 solutions.
However, visual comparisons of the dashed purple line in the volume coordinate frame with the corresponding lines in the C-arm coordinate frame reveal that none of the candidate lengths between $B_2$ and $B_3$ are valid.
Therefore, no solutions would be reported for this source-to-detector distance.}
\label{fig:bb_pose_toy}
\end{figure}
%
%
%
%
\section{Supplementary Results}
\subsection{Intraoperative BB Reconstruction}
\begin{figure}
\begin{center}
\includegraphics[width=\linewidth]{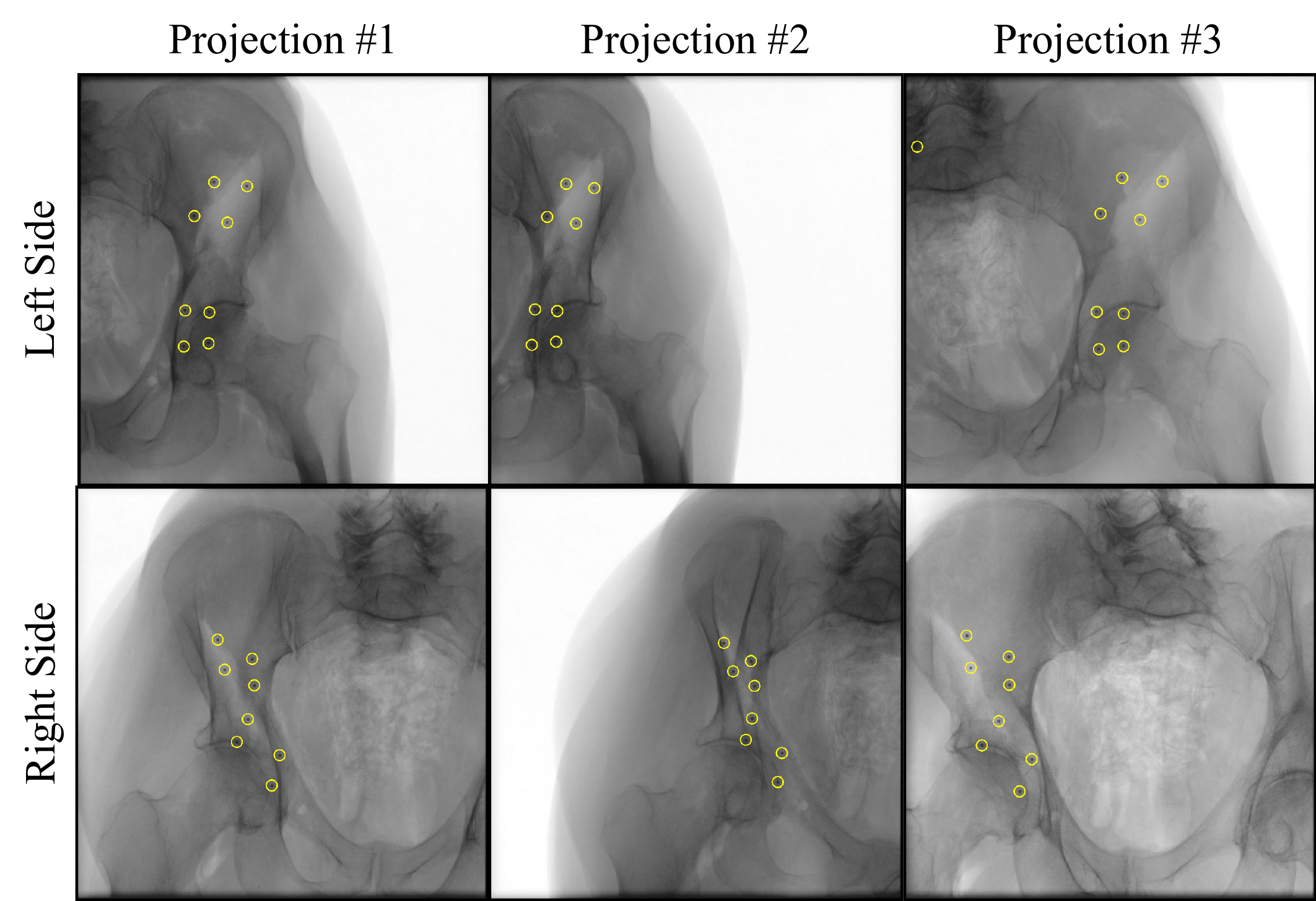}
\end{center}
\caption{The fluoroscopic images used for BB reconstruction in the surgeries for cadaver specimen~1.
The detected BBs are overlaid as yellow circles.
For this specimen, the larger radius parameter passed to the radial symmetry method caused a single false detection on a vertebrae in projection~\#3 of the left side.
This false alarm did not affect the reconstruction.}
\label{fig:spec1_fluoro_precut}
\end{figure}

\begin{figure}
\begin{center}
\includegraphics[width=\linewidth]{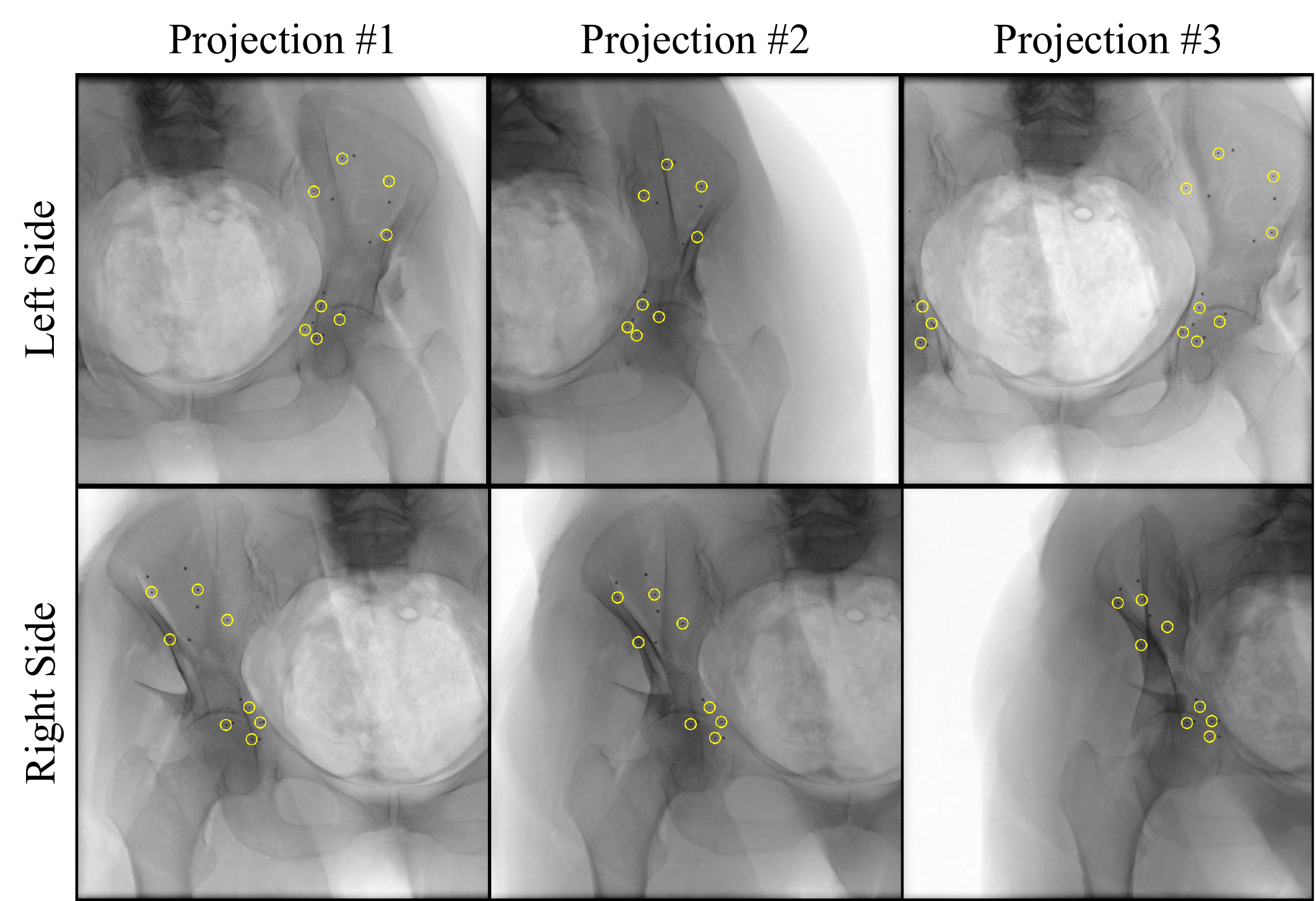}
\end{center}
\caption{The fluoroscopic images used for BB reconstruction in the surgeries for cadaver specimen~2.
The detected BBs are overlaid as yellow circles.
A smaller radius parameter was passed to the radial symmetry method and resulted in no false detections; only injected BBs were detected.
The three contralateral fragment BBs detected in projection \#3 of the left side did not affect the reconstruction.}
\label{fig:spec2_fluoro_precut}
\end{figure}

\begin{figure}
\begin{center}
\includegraphics[width=\linewidth]{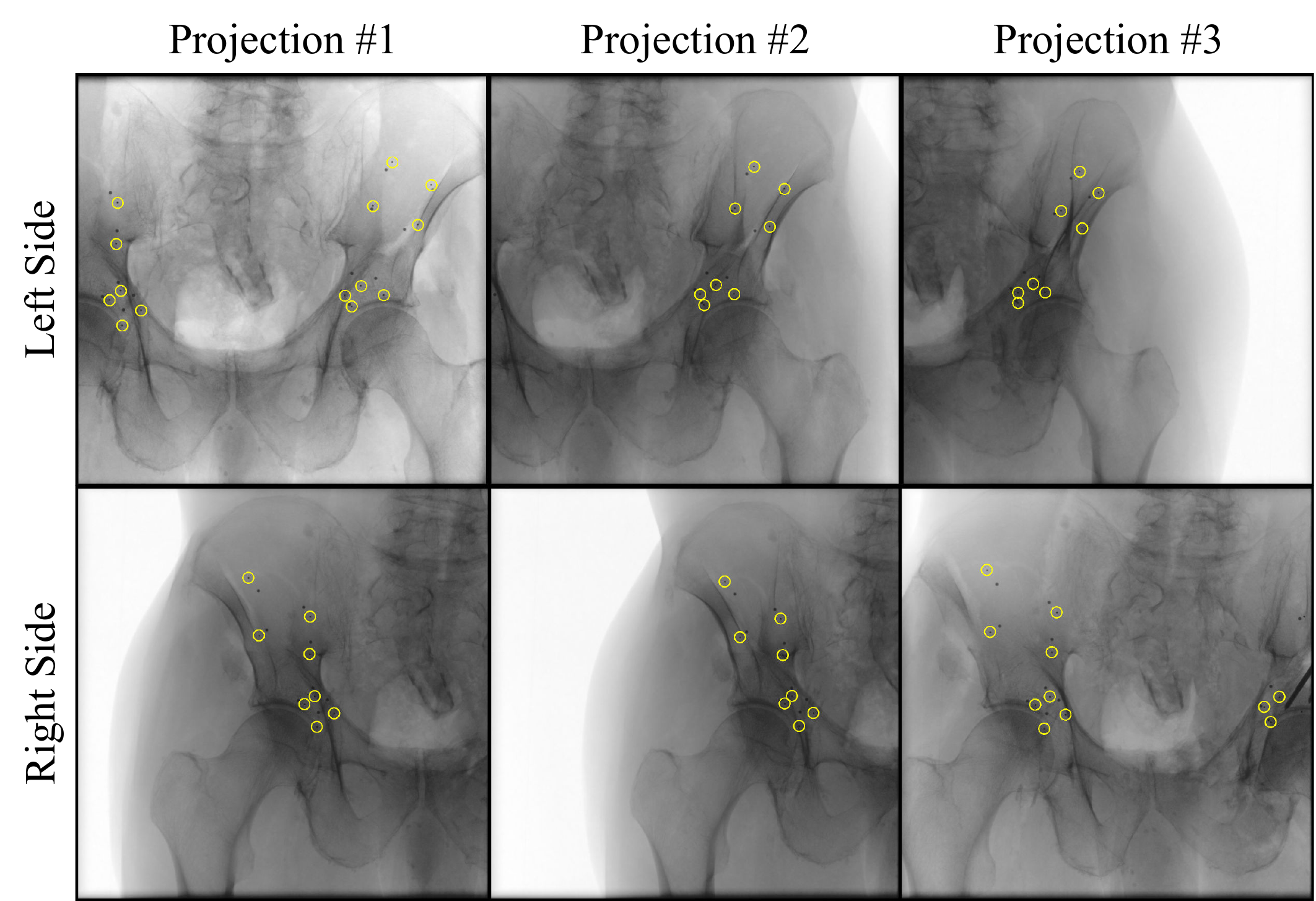}
\end{center}
\caption{The fluoroscopic images used for BB reconstruction in the surgeries for cadaver specimen~3.
The detected BBs are overlaid as yellow circles.
A smaller radius parameter was passed to the radial symmetry method and resulted in no false detections; only injected BBs were detected.
The six contralateral BBs detected in projection \#1 of the left side did not affect the reconstruction.
The three contralateral fragment BBs detected in projection \#3 of the right side, in addition to the K-wire on the contralateral side from a previous PAO, also did not affect the reconstruction.}
\label{fig:spec3_fluoro_precut}
\end{figure}

The fluoroscopy views used for BB reconstruction in the cadaver experiments are shown in figures~\ref{fig:spec1_fluoro_precut},~\ref{fig:spec2_fluoro_precut}, and~\ref{fig:spec3_fluoro_precut}.
Only the smaller injected BBs are detected for the views of specimens~2 and~3; the larger BBs were used for establishing ground truth and not intraoperative reconstruction or pose estimation.
Table~\ref{tab:supp_results_bb_recon_num_dets} shows the number of BBs detected in each fluoroscopic frame used for reconstruction.
A listing of the runtimes associated with BB reconstruction is given in table~\ref{tab:supp_results_bb_recon_times}.
\begin{table}
\caption{The number of BBs detected in each fluoroscopic frame used for reconstruction. There were no missed detections. Numbers greater than 8 indicate that contralateral BBs were detected or there was at least one false detection.}
\label{tab:supp_results_bb_recon_num_dets}
\begin{indented}
\item[]
\begin{tabular}{@{}l r r r}
\br
Surgery    &   View 1  & View 2 & View 3   \\ 
\mr
1 Left       &     8          &     8     &    9         \\
1 Right     &     8          &     8     &    8         \\
2 Left       &     8          &     8     &    11         \\
2 Right     &     8          &     8     &    8         \\
3 Left       &    16          &     8     &    8         \\
3 Right     &     8          &     8     &    11         \\
\br
\end{tabular}
\end{indented}
\end{table}
%
%
\begin{table}
\caption{Runtimes of the different phases of BB reconstruction and the total runtime. All times are listed in seconds.}
\label{tab:supp_results_bb_recon_times}
\begin{indented}
\item[]
\begin{tabular}{@{}l r r r r r r r r}
\br
\multirow{2}{*}{Surgery}    &   \multicolumn{2}{c}{View 1}                                                                                      &  \multicolumn{2}{c}{View 2}                                                                                    & \multicolumn{2}{c}{View 3}                                                                                       & \multirow{2}{25pt}{Recon. Time} & \multirow{2}{25pt}{Total Time}  \\ \cline{2-7}
                                         &  \multicolumn{1}{p{20pt}}{Regi. Time} &  \multicolumn{1}{p{36pt}}{BB Det. Time}   &  \multicolumn{1}{p{20pt}}{Regi. Time} &  \multicolumn{1}{p{36pt}}{BB Det. Time} &  \multicolumn{1}{p{20pt}}{Regi. Time} &  \multicolumn{1}{p{36pt}}{BB Det. Time}   \\
\mr
1 Left       &       4.3        &	   0.1                &	     2.3      &  0.1                 &       1.6       &       0.1               &     0.1                      &  8.6   \\
1 Right     &       4.1        &    0.2                &        1.6      &  0.2                 &       1.7       &       0.2               &     $< 0.1$               &  8.1  \\
2 Left       &       4.9        &    0.2                &        1.7      &  0.2                 &       1.6       &       0.2               &     0.1                      &   8.9 \\
2 Right     &      4.6        &    0.2                 &        1.5      & 0.2                  &       1.5       &       0.2               &     0.1                      &  8.3 \\
3 Left       &       4.1        &    0.2                &        1.5       & 0.2                 &        1.5      &        0.2               &    0.3                      &   8.0 \\
3 Right     &      4.3        &    0.2                 &        1.6       & 0.2                 &        1.6      &       0.2               &     0.1                      &  8.2 \\
\br
\end{tabular}
\end{indented}
\end{table}
%
%
%
\subsection{Intraoperative Pose Estimation}\label{sec:supp_results_intraop_pose_est}
\begin{figure}
\begin{center}
\includegraphics[width=\linewidth]{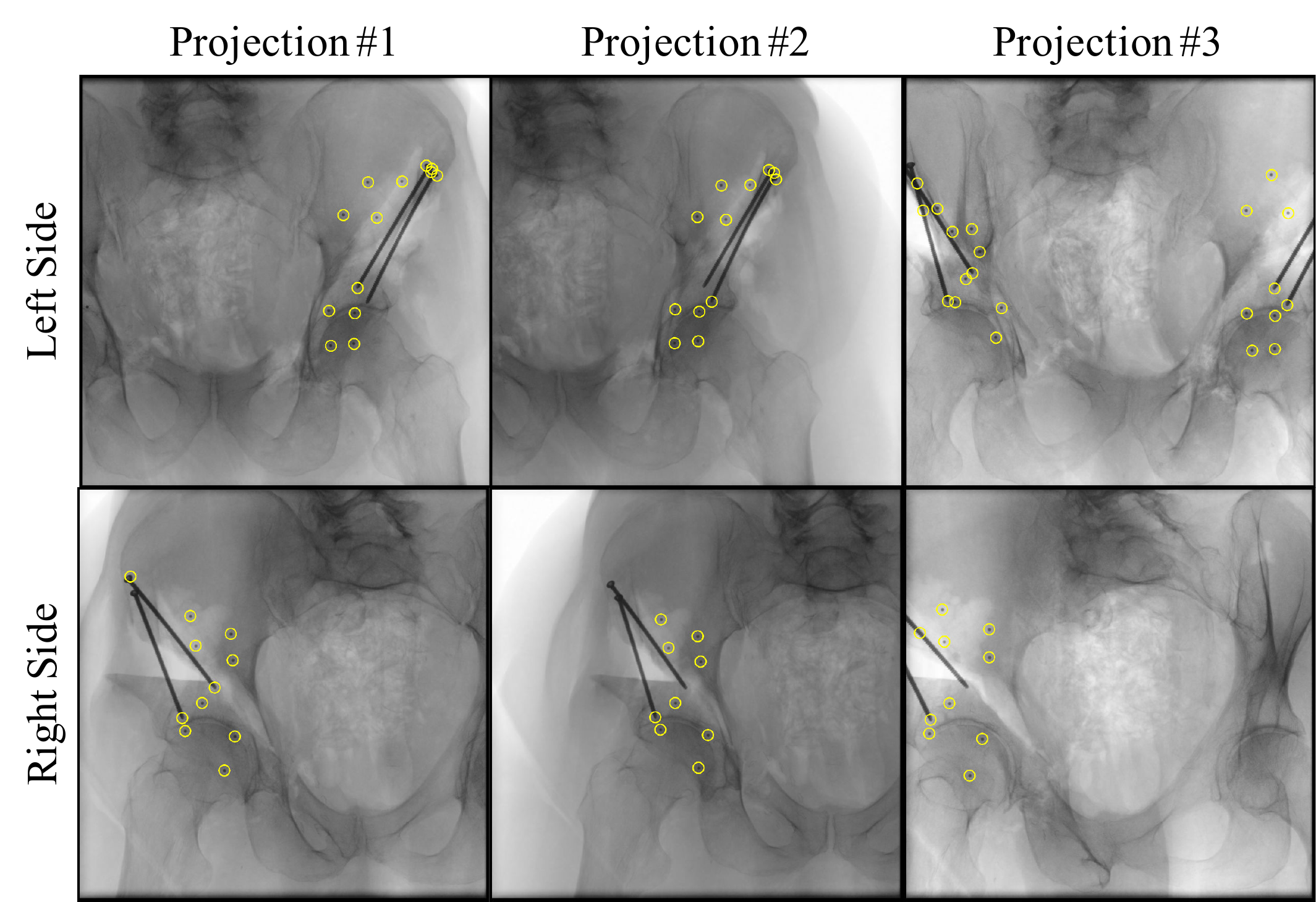}
\end{center}
\caption{The fluoroscopic images used for pose estimation in the surgeries for cadaver specimen~1.
The detected BBs are overlaid as yellow circles.
For this specimen, the larger radius parameter passed to the radial symmetry method caused several false detections on the screws.
Projection 3 on the left side shows an example of an excessive number of detections (21), with 8 detections corresponding to BBs on the contralateral side, 6 false alarms triggered by screws, and the remaining 7 detections corresponding to the desired ipsilateral BBs.}
\label{fig:spec1_fluoro}
\end{figure}

\begin{figure}
\begin{center}
\includegraphics[width=\linewidth]{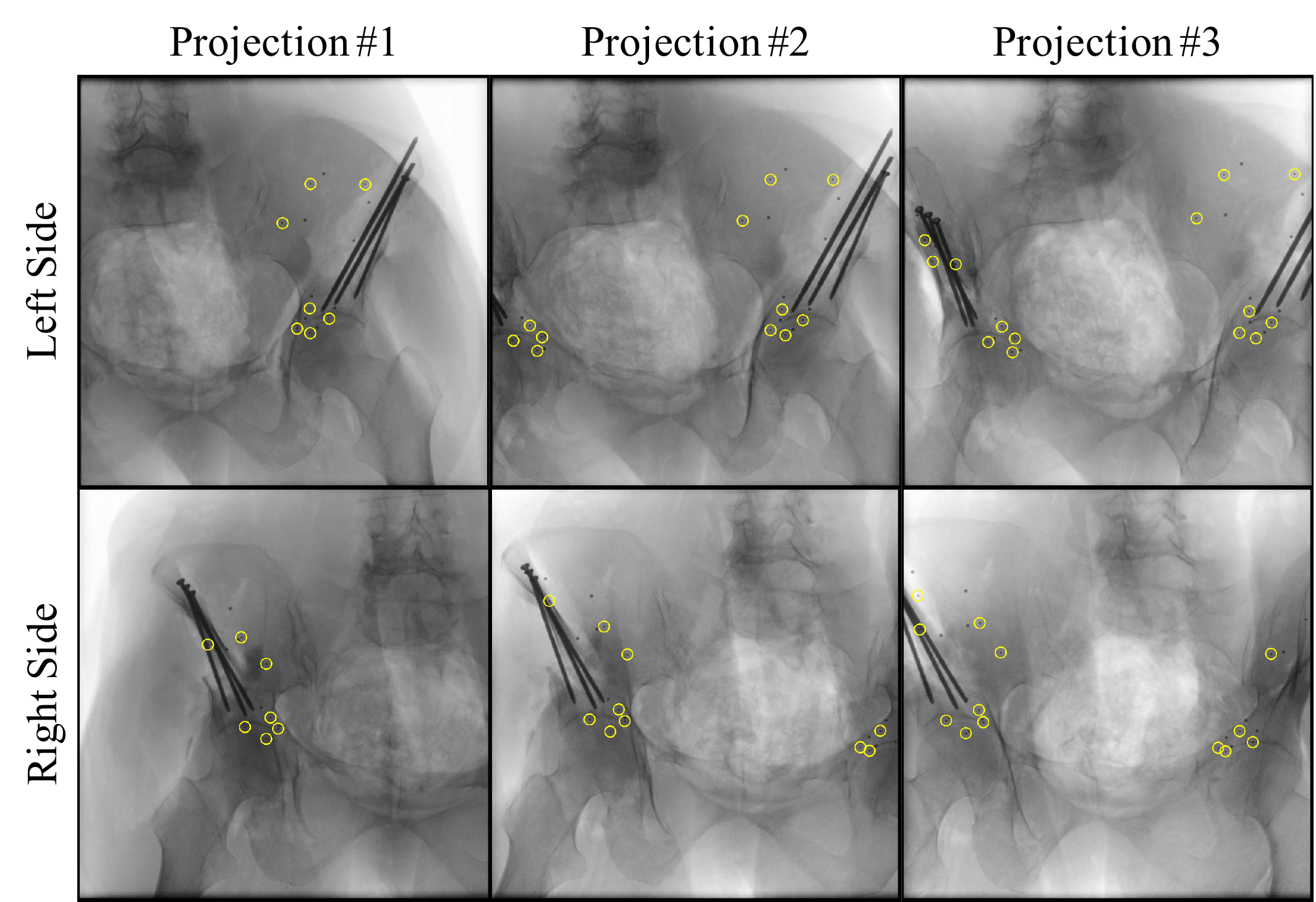}
\end{center}
\caption{The fluoroscopic images used for pose estimation in the surgeries for cadaver specimen~2.
The detected BBs are overlaid as yellow circles.
A smaller radius parameter was passed to the radial symmetry method and resulted in no false detections; only injected BBs were detected.}
\label{fig:spec2_fluoro}
\end{figure}

\begin{figure}
\begin{center}
\includegraphics[width=\linewidth]{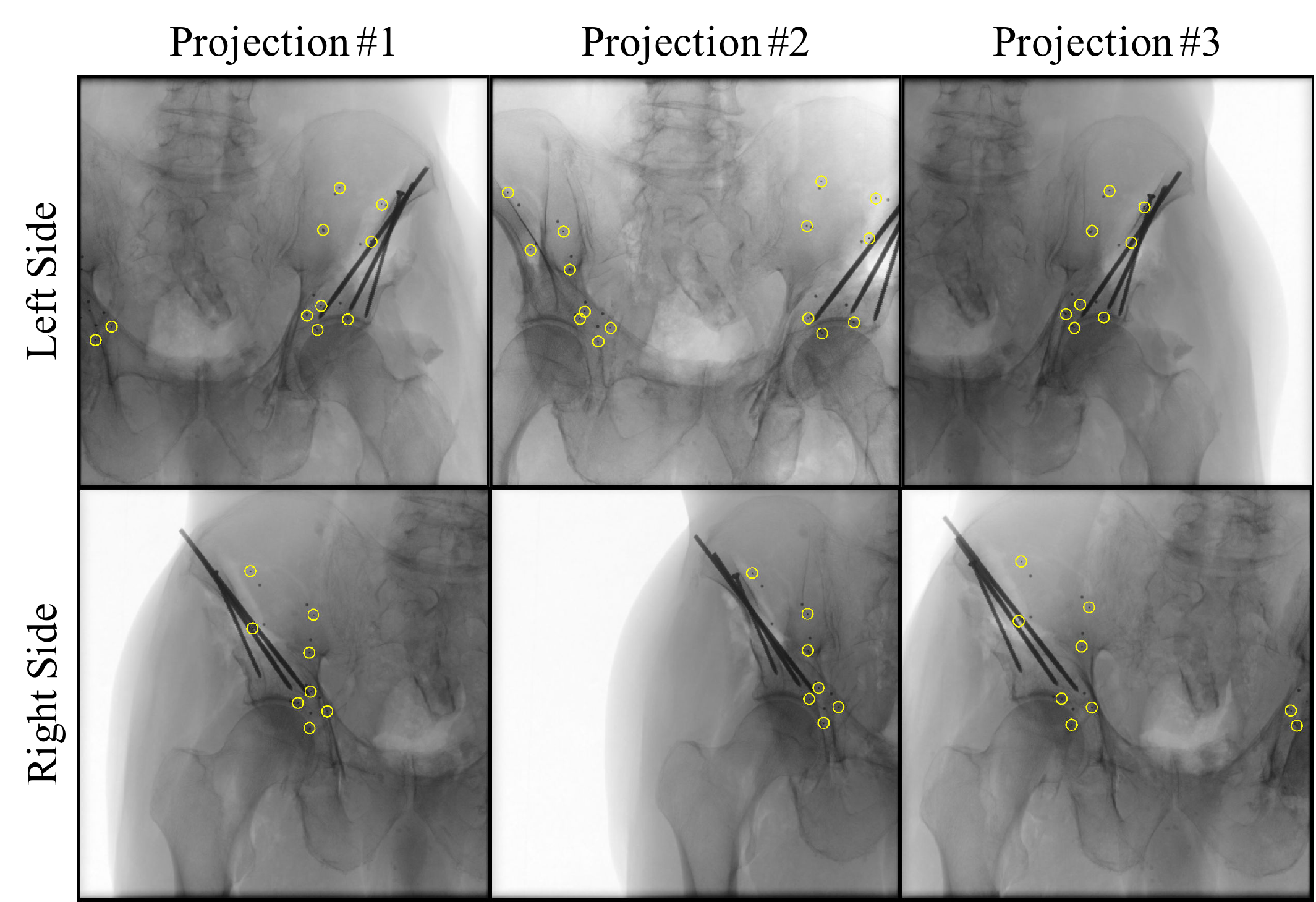}
\end{center}
\caption{The fluoroscopic images used for pose estimation in the surgeries for cadaver specimen~3.
The detected BBs are overlaid as yellow circles.
A smaller radius parameter was passed to the radial symmetry method and resulted in no false detections; only injected BBs were detected.}
\label{fig:spec3_fluoro}
\end{figure}

The fluoroscopy views used for pose estimation in the cadaver experiments are shown in figures~\ref{fig:spec1_fluoro},~\ref{fig:spec2_fluoro}, and~\ref{fig:spec3_fluoro}.
Table~\ref{tab:supp_results_pose_pruning} shows the number of ilium and fragment poses considered at each stage of the pruning algorithms.
Due to the range of source-to-detector distances searched over, the maximum number of poses considered is greater than the maximum number of possible correspondences.
Table~\ref{tab:supp_results_pose_errors} lists the fragment rotation, translation, and lateral center edge (LCE) angle errors for each surgery.
Table~\ref{tab:supp_results_bb_detects} lists the total number of BB detections in each image, along with the number of ilium and fragment BBs matched.
Table~\ref{tab:supp_results_bb_detects_times_pose_est} lists the total number of 2D detections in each image, including false positives, along with the runtimes of the pose estimation pipeline.
\begin{table}
\caption{A summary of the number of pose and correspondence combinations for the ilium and fragment BB constellations during the process of single-view fragment pose estimation for the three different views of each cadaver surgery.
The maximum number of possible combinations are listed, along with the number after each pruning step.
Each of the pose candidates after anatomical pruning for the ilium is used for initialization of the full-pelvis intensity-based 2D/3D registration.
The maximum number of fragment poses and correspondences is lower than that of the ilium, since the ilium correspondences are established first and implausible fragment BB detections are pruned.}
\label{tab:supp_results_pose_pruning}
\begin{indented}
\item[]
\begin{tabular}
{@{}l c r r r r r r} 
\br
\multirow{2}{*}{Surgery} & \multirow{2}{*}{Proj.} & \multicolumn{3}{p{135pt}}{\# Ilium Pose/Correspondence Candidates}                             & \multicolumn{3}{p{135pt}}{\# Frag. Pose/Correspondence Candidates}                                  \\ \cline{3-5} \cline{6-8}
                                       &                                  &  \multicolumn{1}{p{40pt}}{Before Pruning}     & \multicolumn{1}{p{37pt}}{After P3P Pruning} & \multicolumn{1}{p{37pt}}{After Anat. Pruning} & \multicolumn{1}{p{40pt}}{Before Pruning}     & \multicolumn{1}{p{37pt}}{After P3P Pruning} & \multicolumn{1}{p{37pt}}{After Anat. Pruning}                                                             \\
\mr
1 Left & 1 & 885,456 & 2,567 & 155 & 7,920 & 139 & 34 \\
1 Left & 2 & 681,120 & 2,381 & 80 & 7,920 & 157 & 40 \\
1 Left & 3 & 4,117,680 & 6,230 & 168 & 15,840 & 198 & 22 \\
1 Right & 1 & 510,840 & 843 & 24 & 15,840 & 178 & 33 \\
1 Right & 2 & 260,064 & 645 & 17 & 7,920 & 97 & 18 \\
1 Right & 3 & 371,520 & 472 & 14 & 7,920 & 153 & 52 \\
2 Left & 1 & 108,360 & 450 & 12 & 3,168 & 68 & 9 \\
2 Left & 2 & 510,840 & 590 & 10 & 3,168 & 89 & 6 \\
2 Left & 3 & 1,126,944 & 1,204 & 10 & 3,168 & 89 & 10 \\
2 Right & 1 & 108,360 & 1,097 & 70 & 3,168 & 156 & 30 \\
2 Right & 2 & 371,520 & 858 & 23 & 3,168 & 142 & 17 \\
2 Right & 3 & 885,456 & 1,269 & 32 & 3,168 & 106 & 22 \\
3 Left & 1 & 371,520 & 783 & 35 & 3,168 & 122 & 12 \\
3 Left & 2 & 1,408,680 & 1,359 & 48 & 792 & 12 & 4 \\
3 Left & 3 & 173,376 & 958 & 50 & 3,168 & 100 & 12 \\
3 Right & 1 & 173,376 & 1,639 & 129 & 3,168 & 106 & 22 \\
3 Right & 2 & 108,360 & 1,139 & 85 & 3,168 & 108 & 17 \\
3 Right & 3 & 260,064 & 698 & 58 & 792 & 26 & 5 \\
\br
\end{tabular}
\end{indented}
\end{table}
%
%
%
\begin{table}
\caption{A summary of the single-view fragment pose and lateral center edge (LCE) angle errors.
Errors are reported for the three fluoroscopic views taken during each surgery, identified with a cadaver specimen number and operative side, along with the means and standard deviations over all surgeries.
In addition to the rotation and translation pose error magnitudes, a full decomposition of pose errors about anatomical axes is listed.}
\label{tab:supp_results_pose_errors}
\begin{indented}
\item[]
\begin{tabular}{@{}l l l l l l l l l l l} 
\br
\multirow{2}{*}{Surgery}  &  \multirow{2}{*}{Proj.}  &  \multicolumn{4}{c}{Rotation Errors ($\degree$)} &  \multicolumn{4}{c}{Translation Errors (mm)} & \multirow{2}{*}{{LCE ($\degree$)}}  \\ \cline{3-10}
 & & Total & LR & IS & AP & Total & LR & IS & AP & \\
\mr
1 Left&1&1.7&0.3&1.5&0.7&3.1&2.9&0.7&1.0&1.0 \\
1 Left&2&2.5&2.2&0.5&1.1&1.9&1.3&0.1&1.3&0.8 \\
1 Left&3&1.6&0.5&1.3&0.7&3.2&3.1&0.9&0.5&0.9 \\
1 Right&1&1.9&1.9&0.3&0.1&2.0&1.6&1.0&0.5&0.1 \\
1 Right&2&3.3&0.7&2.8&1.5&2.0&0.8&1.2&1.3&1.4 \\
1 Right&3&2.8&2.6&0.1&1.1&2.4&1.5&1.8&0.1&0.9 \\
2 Left&1&3.7&1.4&3.3&0.7&2.3&1.8&1.4&0.4&1.3 \\
2 Left&2&3.3&1.1&3.1&$<0.1$&1.6&0.7&1.0&1.1&0.6 \\
2 Left&3&2.2&0.6&2.1&0.2&1.7&0.3&1.6&0.3&0.8 \\
2 Right&1&2.4&0.9&2.2&0.4&1.1&0.1&0.2&1.1&1.8 \\
2 Right&2&1.7&0.3&1.6&0.4&2.1&0.9&0.3&1.9&1.1 \\
2 Right&3&3.2&0.5&3.1&0.2&1.9&0.6&0.4&1.7&1.7 \\
3 Left&1&0.7&0.4&0.1&0.5&3.0&2.0&1.9&1.3&0.4 \\
3 Left&2&1.3&0.6&0.9&0.6&1.3&1.1&0.7&0.1&1.8 \\
3 Left&3&1.2&1.1&$<0.1$&0.5&2.2&0.6&1.7&1.3&1.8 \\
3 Right&1&2.4&2.3&0.1&0.2&1.6&0.6&0.9&1.2&0.6 \\
3 Right&2&3.0&2.7&1.3&0.3&2.3&0.7&0.7&2.1&0.3 \\
3 Right&3&5.0&2.7&3.6&2.0&1.7&0.8&0.9&1.2&1.4 \\
Mean &  --- & 2.4& 1.3 & 1.6 & 0.6 & 2.1 & 1.2 & 1.0 & 1.0 & 1.0 \\
Std. & --- & 1.0 & 0.9 & 1.2 & 0.5 & 0.6 & 0.8 & 0.5 & 0.6 & 0.5 \\
\br
\end{tabular}
\end{indented}
\end{table}
%
%
%
%
\begin{table}
\caption{A summary of the number of BBs detected in each image and the number matched by the pose estimation process.
The total number of 2D BB detections includes false alarms on screws and BB detections on the contralateral side.
The number of ilium and fragment BB detections, indicate the number of BBs detected from the appropriate constellation; a number less than 4 implies missed-detections.
The number of ilium and fragment BB matches is the number of final correspondences established per constellation for a given set of ilium and fragment poses.}
\label{tab:supp_results_bb_detects}
\begin{indented}
\lineup
\item[]
\begin{tabular}{@{}l c c c c c c}
\br
\multirow{2}{*}{Surgery} & \multirow{2}{*}{Proj.} &  \multirow{2}{40pt}{\# Total Detected} & \multicolumn{2}{c}{Ilium BBs} &  \multicolumn{2}{c}{Fragment BBs} \\ \cline{4-7}
& & &  \# Detected & \# Matched & \# Detected & \# Matched  \\
\mr
1 Left     & 1 & $13$ & 4 & 2 & 4 & 4 \\
1 Left     & 2 & $12$ & 4 & 4 & 4 & 4 \\
1 Left     & 3 & $21$ & 3 & 2 & 4 & 4 \\
1 Right   & 1 & $11$ & 4 & 4 & 4 & 4 \\
1 Right   & 2 & $\09$ & 4 & 2 & 4 & 4 \\
1 Right   & 3 & $10$ & 4 & 4 & 4 & 4 \\
2 Left     & 1 &  $\07$ & 3 & 3 & 4 & 4 \\
2 Left     & 2 & $11$ & 3 & 3 & 4 & 4 \\
2 Left     & 3 & $14$ & 3 & 3 & 4 & 4 \\
2 Right   & 1 &  $\07$ & 3 & 3 & 4 & 4 \\
2 Right   & 2 & $10$ & 3 & 3 & 4 & 4 \\
2 Right   & 3 & $13$ & 4 & 3 & 4 & 4 \\
3 Left     & 1 & $10$ & 4 & 4 & 4 & 4 \\
3 Left     & 2 & $15$ & 4 & 4 & 3 & 3 \\
3 Left     & 3 & $\08$ & 4 & 4 & 4 & 4 \\
3 Right   & 1 & $\08$ & 4 & 3 & 4 & 4 \\
3 Right   & 2 & $\07$ & 3 & 3 & 4 & 4 \\
3 Right   & 3 & $\09$ & 4 & 3 & 3 & 3 \\
\br
\end{tabular}
\end{indented}
\end{table}
%
%
%
%
\begin{table}
\caption{A summary of the number of BBs detected in each image and the runtimes of the pose estimation process.
The total number of 2D BB detections includes false alarms on screws and BB detections on the contralateral side.
All times are in
seconds.}
\label{tab:supp_results_bb_detects_times_pose_est}
\begin{indented}
\lineup
\item[]
\begin{tabular}{@{}l c r r r r}
\br
Surgery & Proj. &  \multicolumn{1}{p{40pt}}{\# Total 2D Dets.} & Det. Time & Pose Est. Time & Total Time  \\ 
\mr
\iftrue
1 Left     & 1 & $13$   & $0.2$ & $0.6$  & $0.8$ \\
1 Left     & 2 & $12$   & $0.2$ &  $0.5$ & $0.7$ \\
1 Left     & 3 & $21$   & $0.2$ &  $0.8$ & $1.0$\\
1 Right   & 1 & $11$   & $0.2$ &  $0.5$ & $0.7$\\
1 Right   & 2 & $\09$  & $0.2$ &  $0.4$ & $0.6$ \\
1 Right   & 3 & $10$   & $0.2$ &  $0.4$ & $0.6$ \\
2 Left     & 1 &  $\07$ & $0.2$ &  $0.4$ & $0.6$\\
2 Left     & 2 & $11$   & $0.2$ &  $0.5$ & $0.7$ \\
2 Left     & 3 & $14$   & $0.2$ &  $0.5$ & $0.7$ \\
2 Right   & 1 &  $\07$ & $0.2$ &  $0.4$ & $0.6$ \\
2 Right   & 2 & $10$  & $0.2$ &   $0.4$ & $0.6$\\
2 Right   & 3 & $13$  & $0.2$ &   $0.4$ & $0.6$\\
3 Left     & 1 & $10$  & $0.2$ &   $0.4$ & $0.6$\\
3 Left     & 2 & $15$  & $0.2$ &   $0.5$ & $0.7$\\
3 Left     & 3 & $\08$ & $0.2$ &   $0.4$ & $0.6$\\
3 Right   & 1 & $\08$ & $0.2$ &   $0.6$ & $0.8$\\
3 Right   & 2 & $\07$ & $0.2$ &   $1.2$ & $1.4$\\
3 Right   & 3 & $\09$ & $0.2$ &   $0.4$ & $0.6$\\
\else
1 Left     & 1 & $13$   & $185$ & $630$  & $814$ \\
1 Left     & 2 & $12$   & $183$ &  $508$ & $691$ \\
1 Left     & 3 & $21$   & $183$ &  $794$ & $978$\\
1 Right   & 1 & $11$   & $186$ &  $456$ & $642$\\
1 Right   & 2 & $\09$  & $186$ &  $384$ & $570$ \\
1 Right   & 3 & $10$   & $186$ &  $406$ & $592$ \\
2 Left     & 1 &  $\07$ & $226$ &  $354$ & $580$\\
2 Left     & 2 & $11$   & $226$ &  $477$ & $703$ \\
2 Left     & 3 & $14$   & $226$ &  $474$ & $700$ \\
2 Right   & 1 &  $\07$ & $223$ &  $434$ & $657$ \\
2 Right   & 2 & $10$  & $215$ &   $424$ & $638$\\
2 Right   & 3 & $13$  & $215$ &   $439$ & $654$\\
3 Left     & 1 & $10$  & $232$ &   $364$ & $596$\\
3 Left     & 2 & $15$  & $230$ &   $464$ & $694$\\
3 Left     & 3 & $\08$ & $230$ &   $364$ & $594$\\
3 Right   & 1 & $\08$ & $230$ &   $558$ & $788$\\
3 Right   & 2 & $\07$ & $230$ &   $1,168$ & $1,399$\\
3 Right   & 3 & $\09$ & $231$ &   $372$ & $603$\\
\fi
\br
\end{tabular}
\end{indented}
\end{table}
\end{document}